\documentclass[lettersize,journal]{IEEEtran}

\usepackage{amssymb}
\usepackage{lipsum}
\usepackage{balance}

\usepackage{amsmath,amsfonts}
\usepackage{algorithmic}
\usepackage{algorithm}
\usepackage{array}
\usepackage[caption=false,font=normalsize,labelfont=sf,textfont=sf]{subfig}
\usepackage{textcomp}
\usepackage{stfloats}
\usepackage{url}
\usepackage{verbatim}
\usepackage{graphicx}
\usepackage{cite}
\usepackage{listings}
\usepackage{color}
\usepackage{amsmath}
\usepackage[T1]{fontenc}
\usepackage{hyperref}

\usepackage{listings}
\usepackage{color}
\usepackage{amsmath}
\usepackage{booktabs,chemformula}
\usepackage{graphicx}
\usepackage{placeins}
\usepackage{multicol}
\usepackage{pythonhighlight}

\definecolor{dkgreen}{rgb}{0,0.6,0}
\definecolor{gray}{rgb}{0.5,0.5,0.5}
\definecolor{mauve}{rgb}{0.58,0,0.82}
\usepackage[font=normalsize,labelfont=bf]{caption}
\usepackage{epstopdf}
\AppendGraphicsExtensions{.tif}

\lstdefinelanguage{cypher}
{
	morekeywords={
		MATCH, OPTIONAL, WHERE, NOT, AND, OR, XOR, RETURN, DISTINCT, ORDER, BY, ASC, ASCENDING, DESC, DESCENDING, UNWIND, AS, UNION, WITH, ALL, CREATE, DELETE, DETACH, REMOVE, SET, MERGE, SET, SKIP, LIMIT, IN, CASE, WHEN, THEN, ELSE, END,
		INDEX, DROP, UNIQUE, CONSTRAINT, EXPLAIN, PROFILE, START,
	}
}

\newcommand{\mycdots}{\cdot\!\cdot\!\cdot}
\newcommand{\customsize}{\fontsize{8.2}{10}\selectfont}
\lstdefinestyle{json}{
    basicstyle=\customsize\ttfamily, 
    breaklines=true, 
    postbreak=\mbox{\textcolor{red}{$\hookrightarrow$}\space}, 
    frame=single, 
    rulecolor=\color{black}, 
    backgroundcolor=\color{white}, 
    captionpos=b, 
    showstringspaces=false, 
    tabsize=5, 
    language=Java, 
    morekeywords={text, task_plan, is_navigation_required, actions, action, subtasks, detect_object, verify, localize, grasp, strategy, grip_parameters, contact_points, applied_force, approach_angle, put_in, placement_constraints, max_force, surface_stability, performance_metrics, success_criteria, estimated_time, safety_checks, criteria, name, condition, result, final_decision},
    keywordstyle=\color{blue}, 
    stringstyle=\color{red}, 
    commentstyle=\color{green}, 
    stepnumber=1, 
    numbersep=5pt, 
}

\begin{document}

\title{Safety Control of Service Robots with LLMs and Embodied Knowledge Graphs}

\author{
    \IEEEauthorblockN{Yong Qi\IEEEauthorrefmark{1}, Gabriel Kyebambo\IEEEauthorrefmark{0}, Siyuan Xie\IEEEauthorrefmark{0}, Wei Shen\IEEEauthorrefmark{0}, Shenghui Wang\IEEEauthorrefmark{0}, Bitao Xie\IEEEauthorrefmark{0}, Bin He\IEEEauthorrefmark{0}, Zhipeng Wang\IEEEauthorrefmark{0}, Shuo Jiang\IEEEauthorrefmark{0}}\\
    \IEEEauthorblockA{School of Electronic Information and Artificial Intelligence, Shaanxi University of Science and Technology, Xian, Shaanxi Province, 710026, China\\
  }
}

\maketitle

\begin{abstract}
	Safety concerns in service robotics necessitate robust mechanisms to prevent harm to humans or property, particularly in dynamic environments where human-robot interactions are frequent. While recent advancements integrate Knowledge Graphs (KGs) with Large Language Models (LLMs), significant challenges remain in ensuring consistent safety. These challenges include (1) the potential for misunderstanding natural language commands, (2) conflicts between user commands and environmental constraints, and (3) errors in the robot's perception system that could lead to unsafe actions. To address these issues, this work proposes a comprehensive framework combining LLMs, Embodied Robotic Control Prompts (ERCPs), and an Embodied Knowledge Graph (EKG). The EKG framework incorporates Graph Attention Networks (GATs) to prioritize critical nodes and edges during validation, ensuring real-time awareness of environmental changes, and Hamiltonian Paths to enforce the correct sequence of actions. ERCPs guide LLMs in generating safe task plans by iteratively refining ambiguous human instructions, combining predefined action primitives, and incorporating contextual clarifications. This integration not only enhances safety but also improves the robot's ability to handle complex tasks in dynamic environments. 
\end{abstract}

\begin{IEEEkeywords}
	Large Language Models, Embodied Robotic Control Prompts, Embodied Knowledge Graphs, Graph Attention Networks, Hamiltonian Paths, Service Robots, Safety Framework, Task Plan Validation, Human-Robot Interactions, Real-Time Validation, Perception Error Recovery
\end{IEEEkeywords}

\section{Introduction}
Service robots \cite{gonzalez2021service,paluch2020service} are increasingly deployed in shared living and working spaces such as homes, hospitals, and offices, where they interact closely with humans \cite{holland2021service, wirtz2021service,qiu2020enhancing}. In these environments, ensuring the safety of human-robot interactions is paramount. For example, in a home setting, a robot must avoid dropping fragile objects or colliding with humans, while in a hospital, a robot delivering medication must ensure that sensitive items are handled and delivered without damage or misplacement. Despite their potential, the integration of service robots into human-centric environments raises significant safety concerns \cite{ceccarelli2011problems,kortner2016ethical,guiochet2017safety}, necessitating robust mechanisms to ensure operational reliability and task safety \cite{villani2018survey,tojib2022service}. Safety in service robots involves preventing accidents, injuries, or property damage during human-robot interactions. A critical aspect is the robot's ability to halt or recover from unsafe tasks \cite{huang2020safe}, ensuring actions remain within predefined safety limits \cite{vasic2013safety}.

Despite advancements in robotic control and artificial intelligence, service robots face persistent challenges in \textbf{real-time validation}, \textbf{contextual knowledge representation}, and \textbf{logical reasoning} \cite{ortenzi2021object}. These limitations hinder their safe and efficient operation in dynamic environments, limiting real-world adoption \cite{soroka2012challenges}. For example, misidentifying object properties (e.g., weight or fragility) can lead to unsafe actions, such as dropping objects or applying incorrect force \cite{torras2016service,matsuzaki2016autonomy}. Addressing these challenges is essential for reliable human-robot collaboration \cite{hu2024deploying,huang2023voxposer}.

To enhance safety, approaches leveraging \textbf{Large Language Models (LLMs)} \cite{zhang2023large,zhang2021patterns} and \textbf{Knowledge Graphs (KGs)} \cite{lim2010ontology,hanheide2017robot} have been explored. LLMs enable robots to interpret human instructions and generate task plans but lack real-time environmental awareness \cite{wu2024safety,yang2024safety}. KGs provide structured knowledge for decision-making but struggle with natural language nuances and unstructured data \cite{saxena2014robobrain,lecue2020role}. Recent efforts integrate LLMs and KGs to address these limitations \cite{stark2023dobby,liu2024joint}, combining natural language understanding with structured knowledge.

This work introduces \textbf{Embodied Robotic Control Prompts (ERCPs)} and an \textbf{Embodied Knowledge Graph (EKG)} framework incorporating \textbf{Graph Attention Networks (GATs)} and \textbf{Hamiltonian Paths}. ERCPs guide LLMs in generating safe task plans by combining human instructions, predefined action primitives, and contextual clarifications. The EKG framework, enhanced with GATs, prioritizes critical nodes and edges during validation, while Hamiltonian Paths enforce the correct sequence of actions. This integration addresses challenges in real-time validation, contextual knowledge representation, and logical reasoning, enabling safer and more reliable service robots in dynamic environments.

\subsection{Problem Statement}
Service robots operating in shared spaces (e.g., homes, hospitals, offices) face critical safety challenges when interpreting natural language commands in dynamic environments. These challenges manifest in three key scenarios:

\begin{itemize}
	\item \textbf{Misunderstanding Natural Language Commands}: Ambiguous instructions (e.g., \textit{"Get me a drink"}) lack details like object type, location, or delivery destination. Such ambiguity risks unsafe actions, such as fetching a hot beverage without specifying handling constraints. 
	
	\item \textbf{Conflict Between Commands and Environmental Constraints}: User instructions may conflict with real-world conditions. For example, a hospital robot ordered to deliver medication might encounter blocked paths or improperly stored items, requiring immediate plan adjustment to avoid harm.
	
	\item \textbf{Perception System Errors}: Sensor inaccuracies (e.g., misdetecting a fragile vase as sturdy) can lead to unsafe grasps or collisions, especially in cluttered spaces like offices.
\end{itemize}
\begin{figure*}[t]
	\centering
	\includegraphics[scale=0.30]{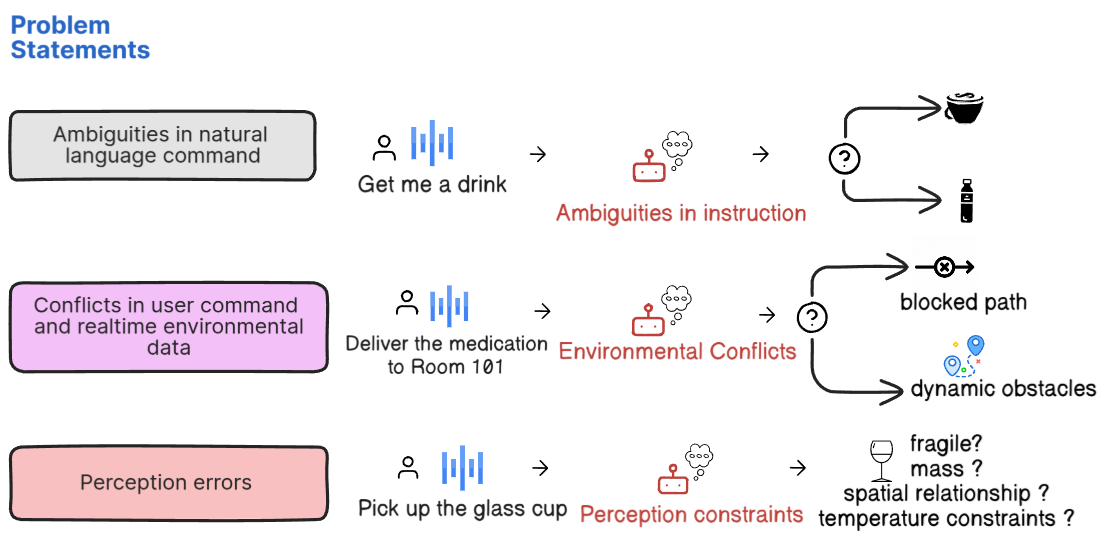} 
	\caption{\textbf{Problem Statement:} The diagram highlights challenges in service robot safety—ambiguities in natural language commands, conflicts between user commands and environmental constraints, and perception errors.}
	\label{fig:ekg_structure}
\end{figure*}

\noindent These challenges translate to three technical sub-problems:
\begin{itemize}
	\item \textbf{Real-Time Validation}: Task plans must adapt dynamically to environmental changes (e.g., object fragility updates) to prevent unsafe actions.
	
	\item \textbf{Fast Graph Retrieval}: Rapid access to environmental data (e.g., obstacle positions) is critical for real-time safety compliance.
	
	\item \textbf{Logical Reasoning}: Task plans must align with physical constraints (e.g., force limits, spatial relationships) even in dynamic scenarios.
\end{itemize}

\noindent Our framework addresses these sub-problems through \textbf{Embodied Robotic Control Prompts (ERCPs)} for resolving ambiguities, an \textbf{Embodied Knowledge Graph (EKG)} for real-time validation, and \textbf{Graph Attention Networks (GATs)} with \textbf{Hamiltonian Paths} to prioritize critical constraints and optimize task sequencing.

\section{Related Work}
\label{sec:related_work}

The use of Large Language Models (LLMs) and Knowledge Graphs (KGs) in robotics has gained significant attention, particularly in safety-critical applications. This section reviews state-of-the-art approaches in LLMs, KGs, and their integration, highlighting their strengths, limitations, and the gaps addressed by the proposed framework.

\subsection{Prompt Engineering}
The development of custom prompt templates for guiding Large Language Models (LLMs) in downstream tasks represents a significant advancement in computational linguistics. Cloze prompts \cite{jia2022visual} and prefix prompts \cite{chen2021lightner} exemplify structured methods that enhance LLMs' task-specific performance. Studies such as \cite{liu2023pretrain} and \cite{jiang2022promptmaker} demonstrate how precise, context-relevant interactions improve LLM efficiency and accuracy across applications, including conversational AI and decision-making. Recent work by \cite{marvin2023prompt} highlights the importance of tailored prompt engineering for optimizing LLM performance in specific domains, such as robotics.

Ding et al. \cite{ding2021openprompt} provide a comprehensive overview of prompt design techniques, categorizing them into manual and automatic methods. While manual methods offer bespoke customization, automatic methods, particularly those generating cloze prompts, are favored for their scalability and standardization. White et al. \cite{white2023prompt} expand on this by introducing a catalog of prompt patterns adaptable for robotics applications, improving task planning and execution.

The experimental study ``ChatGPT for Robotics'' \cite{vemprala2023chatgpt} showcases the integration of prompt engineering with robust function libraries, enhancing adaptability across diverse robotics applications. Recent advances in LLMs have revolutionized robotics, with studies like \cite{bode2024comparison} comparing prompt engineering techniques and \cite{kim2024survey} providing guidelines for integrating LLMs across robotic functionalities. Shirasaka et al. \cite{shirasaka2023self} introduce self-recovery prompting, enabling service robots to autonomously recover from errors, while Zhou et al. \cite{zhou2022conditional} explore conditional prompt learning for vision-language models, enhancing multimodal capabilities.

Building on these works, we propose Embodied Robotic Control Prompts (ERCPs). ERCPs are prompt templates composed of natural language commands \( c \), text clarifications, and defined action primitives (APIs) to enhance safe task plan generation by LLMs.

\subsection{Large Language Models in Robotics}
Large Language Models (LLMs) have emerged as transformative tools in robotics, enabling \textbf{multi-modal semantic understanding} and enhancing robots' ability to interpret and interact with their environments. Recent advancements in fine-tuning and pre-training methods have improved LLMs' ability to process and integrate information across multiple modalities \cite{kojima2022large}.

A key strength of LLMs lies in their ability to leverage \textbf{text descriptions} to extract semantic information, facilitating a deeper understanding of environmental contexts. This capability enables robots to execute tasks through \textbf{action primitives (APIs)}, ensuring more intuitive and adaptive interactions \cite{huang2022inner}. For instance, Wu et al. \cite{wu2023tidybot} developed a framework integrating language-based planning with perception, enabling robots to adapt to user preferences with minimal prior interactions, particularly in \textbf{user-centric applications}.

LLMs have also been applied to complex tasks such as \textbf{multi-object rearrangement}, where Ding et al. \cite{ding2023task} demonstrated their utility in enhancing robotic motion planning and execution. Similarly, systems like \textbf{CodeBotler} \cite{hu2023deploying} and \textbf{LLM-Grounder} \cite{yang2023llm} highlight the versatility of LLMs in programming service robots and improving \textbf{3D visual grounding}, respectively.

Further contributions include \textbf{ProgPrompt} \cite{singh2023progprompt} and \textbf{Code as Policies} \cite{liang2023codepolicies}, which adapt LLMs for generating dynamic task plans and policy codes from natural language instructions. These frameworks enable robots to exhibit agile and adaptable behaviors. However, challenges such as \textbf{linguistic ambiguities} and \textbf{hallucinations} in LLM outputs can hinder task execution. Wang et al.~\cite{wang2024llm} address these issues with a constrained LLM prompt scheme that generates executable action sequences and incorporates an exception handling module to mitigate hallucinations. Additionally, REFLECT~\cite{reflect2023} introduces a reflection-based evaluation mechanism that helps reduce hallucinations in LLM-generated reasoning chains, improving task reliability.

\subsection{Knowledge Graphs in Robotics}
Knowledge Graphs (KGs) play a central role in advancing robotics, improving robot manipulation, autonomous task planning, and human-robot interaction. Miao et al. \cite{miao2023semantic} introduced a framework for \textbf{Semantic Representation of Robot Manipulation with Knowledge Graphs}, employing a multi-layered semantic model to enhance precision in manipulation and task execution, critical for real-world applications.

Bai et al. \cite{bai2024dynamic} proposed a dynamic KG approach for distributed self-driving laboratories, integrating various components into a cohesive digital twin. This method adapts to changing research objectives and facilitates resource integration, demonstrating the importance of dynamic KGs in real-time control and mitigating geographical constraints.

In human-robot collaboration, Ding et al. \cite{ding2019robotic} used KGs to enhance complex tasks like disassembly, equipping robots with improved knowledge and reasoning capabilities. Peng et al. \cite{peng2023knowledge} further explored challenges in KGs, such as addressing long-tail knowledge, biases, and the need for explainability and interpretability.

Context representation is crucial for robotic systems. While ontologies provide structured knowledge, they lack efficient visualization and querying. Dimitropoulos et al. \cite{dimitropoulos2024ontology} proposed a hybrid approach, converting ontologies into KGs to combine their strengths, offering better consistency and accessibility for robotic applications. However, KGs lack logical consistency checking, a feature of ontologies.

Recent advancements in KG frameworks, such as \textbf{OpenKE} \cite{han2017openke} and \textbf{PyTorch-BigGraph} \cite{lerer2019pytorchbiggraph}, have enhanced the scalability and efficiency of knowledge representation, enabling their application in complex robotic systems.

\subsection{Integrating LLMs and EKGs}
The integration of Large Language Models (LLMs) and Knowledge Graphs (KGs) aims to combine their strengths, with LLMs excelling in language prediction and KGs providing structured, factual knowledge~\cite{yang2023chatgpt}. However, LLMs alone lack grounding in facts, while KGs struggle with natural language nuances. Alam et al.~\cite{alam2022language} improve KG embeddings for graph completion, addressing issues like neglecting textual data. Pan et al.~\cite{pan2024unifying} explore strategies for integrating LLMs with KGs to enhance LLM inference. Despite progress, challenges like frequent updates and generalization to unseen data persist~\cite{mccoy2019right}. SAFETY CHIP~\cite{safetychip2023} adds a layer of safety to LLM-based decisions in robotics.

Dynamic Knowledge Fusion methods, such as the Two-Tower Architecture~\cite{wang2019improving}, process text and KG data separately, while advanced models like KagNet~\cite{lin2019kagnet} and MHGRN~\cite{feng2020scalable} incorporate KG data into the text, and GreaseLM~\cite{zhang2022greaselm} and QA-GNN~\cite{yasunaga2021qa} introduce deep interaction mechanisms.

Our approach integrates LLMs with Embodied Robotic Control Prompts (ERCPs) and an Embodied Knowledge Graph (EKG), utilizing Neo4j \cite{neo4j2023} and Cypher queries \cite{francis2018cypher} for efficient data management. A Graph Attention Network (GAT) model computes attention weights for critical nodes in the EKG, guiding task plan validation. Hamiltonian Paths ensure the correct sequence of critical nodes during execution.

This integration is formulated as a \textbf{multi-stage optimization problem}:
\[
S = \arg\max_{\mathcal{P} \in \mathcal{T}} \mathcal{F}(\mathcal{P}, K_{\text{EKG}}, P_{\text{ERCP}}),
\]
where:
\begin{itemize}
	\item \( \mathcal{P} \): A candidate task plan,
	\item \( \mathcal{T} \): The space of possible plans,
	\item \( K_{\text{EKG}} \): Knowledge from the EKG, represented as a graph \( G = (\mathcal{V}, \mathcal{E}) \), with entities \( \mathcal{V} \) and relationships \( \mathcal{E} \),
	\item \( P_{\text{ERCP}} \): A combination of user instructions, action primitives, and clarifications, formalized as \( P_{\text{ERCP}} = \{c, \mathcal{A}, \mathcal{T}\} \), where \( c \) is the user command, \( \mathcal{A} \) is the action set, and \( \mathcal{T} \) represents textual clarifications.
\end{itemize}

The \textbf{objective function} \( \mathcal{F} \) evaluates the feasibility and safety of the plan:
\[
\mathcal{F}(\mathcal{P}, K_{\text{EKG}}, P_{\text{ERCP}}) = \sum_{a \in \mathcal{P}} \mathcal{V}(a, K_{\text{EKG}}, P_{\text{ERCP}}),
\]
where \( \mathcal{V}(a, K_{\text{EKG}}, P_{\text{ERCP}}) \) is a validation function that checks if action \( a \) meets safety and feasibility criteria:
\[
\mathcal{V}(a, K_{\text{EKG}}, P_{\text{ERCP}}) =
\begin{cases}
	1 & \text{if } a \text{ is safe and feasible}, \\
	0 & \text{otherwise}.
\end{cases}
\]

To ensure the correct sequence of visits, a \textbf{Hamiltonian Path constraint} is added:
\[
\mathcal{H}(\mathcal{P}, K_{\text{EKG}}) = 
\begin{cases}
	1 & \text{if } \mathcal{P} \text{ visits all critical nodes in order}, \\
	0 & \text{otherwise}.
\end{cases}
\]

The validated task plan \( S \) is determined by solving the constrained optimization problem:
\[
S = \arg\max_{\mathcal{P} \in \mathcal{T}} \mathcal{F}(\mathcal{P}, K_{\text{EKG}}, P_{\text{ERCP}})
\quad \text{subject to} \quad \mathcal{H}(\mathcal{P}, K_{\text{EKG}}) = 1.
\]

\begin{table}[ht]
	\centering
	\caption{Comparative Analysis of Key Features: Knowledge Graphs as a Knowledge Base, Operation in Unseen Environments, Plan Visualization and Correction, Plan Optimization, and Safety Evaluation — Our Approach vs. Related Works}
	\small 
	\renewcommand{\arraystretch}{1.6} 
	\resizebox{\columnwidth}{!}{
		\begin{tabular}{p{0.18\columnwidth} p{0.12\columnwidth} p{0.12\columnwidth} p{0.12\columnwidth} p{0.12\columnwidth} p{0.12\columnwidth} p{0.12\columnwidth}}
			\hline
			& \textbf{KG as KB} & \textbf{Unseen Env.} & \textbf{Plan Vis.} & \textbf{Plan Corr.} & \textbf{Plan Opt.} & \textbf{Safety Eval.} \\
			\hline
			CodeBotler \cite{hu2023deploying} & x & -- & \checkmark & \checkmark & \checkmark & x  \\
			ProgPrompt \cite{singh2023progprompt} & x & \checkmark & \checkmark & \checkmark & x & x \\
			Code as Policies \cite{liang2023codepolicies} & x & \checkmark & \checkmark & \checkmark & \checkmark & x \\
			REFLECT \cite{reflect2023} & x & -- & \checkmark & \checkmark & x & x \\
			ToT \cite{yang2023chatgpt} & x & \checkmark & \checkmark & \checkmark & \checkmark & x \\
			Safety Chip \cite{safetychip2023} & x & -- & x & x & x & \checkmark \\
			\hline
			Ours & \checkmark & \checkmark & \checkmark & \checkmark & \checkmark & \checkmark \\
			\hline
		\end{tabular}
	}
	\label{tab:key_features}
\end{table}

\begin{figure*}[t]
	\centering
	\includegraphics[scale=0.28]{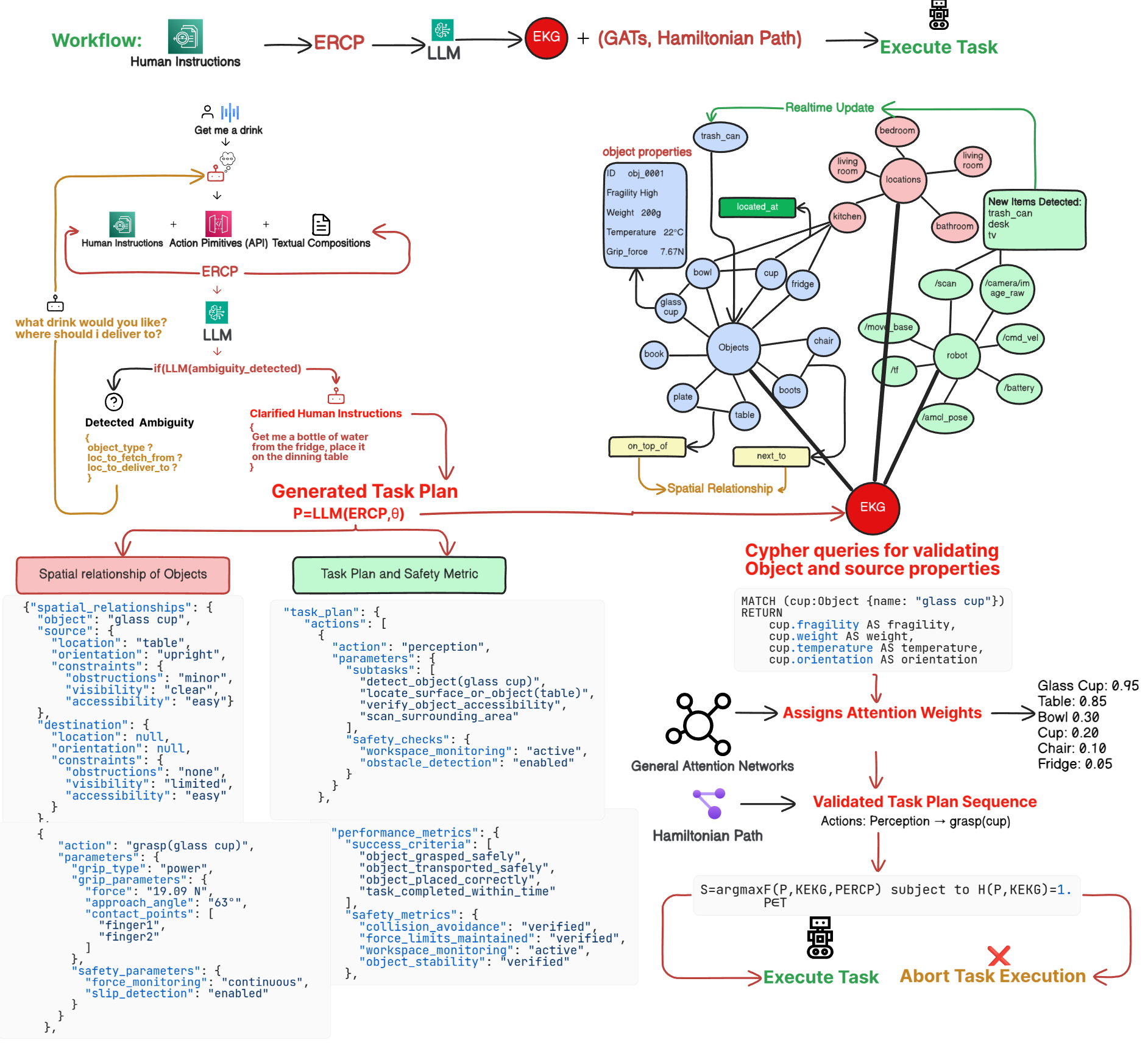} 
	\caption{\textbf{Integrated Framework for Service Robot Safety and Control:} The diagram illustrates the interaction between the \textbf{Embodied Robotic Control Prompt (ERCP)}, \textbf{Large Language Model (LLM)}, \textbf{Embodied Knowledge Graph (EKG)}, \textbf{Graph Attention Networks (GATs)}, and \textbf{Hamiltonian Paths}. The ERCP refines ambiguous user instructions, the LLM generates task plans, the EKG validates environmental and object constraints, GATs prioritize safety-critical nodes, and Hamiltonian Paths ensure the correct sequence of actions for safe and efficient task execution.}
	\label{fig:overall_framework}
\end{figure*}

\section{Methodology}

This work presents an integrated framework for enhancing safety and control in service robots through four main components: (1) the Embodied Robotic Control Framework, which manages high-level task planning and execution; (2) the Knowledge Representation system, built around an Embodied Knowledge Graph (EKG); (3) the Task Planning and Execution module, which implements a Markov Decision Process (MDP) framework; and (4) the Safety and Validation layer, which ensures safe operation through continuous monitoring and constraint verification.

\subsection{Problem Abstraction}
The core challenge in safe human-robot interaction is ensuring that service robots execute natural language commands safely in dynamic environments. This requires addressing three interconnected sub-problems:
\begin{itemize}
	\item \textbf{Real-Time Validation}: Task plans must adapt to environmental changes (e.g., moving obstacles, altered object properties) during execution to prevent unsafe actions.
	\item \textbf{Fast Graph Retrieval}: Environmental data (e.g., object positions, fragility) must be retrieved and updated rapidly to avoid latency-induced safety risks.
	\item \textbf{Logical Consistency}: Task plans must adhere to physical constraints (e.g., force limits, spatial relationships) even as the environment evolves.
\end{itemize}  

These challenges stem from the inherent limitations of LLMs in grounding language to real-world physics and the computational overhead of traditional KGs. Our work addresses these gaps by integrating \textbf{Embodied Robotic Control Prompts (ERCPs)} for instruction refinement, an \textbf{Embodied Knowledge Graph (EKG)} for structured environmental representation, and algorithmic optimizations (\textbf{GATs}, \textbf{Hamiltonian Paths}) to prioritize safety-critical constraints.  

\subsection{Embodied Robotic Control Prompt (ERCP)}
The \textbf{Embodied Robotic Control Prompt (ERCP)} framework provides a structured mechanism for guiding Large Language Models (LLMs) to clarify ambiguous or incomplete human instructions through iterative refinement. The ERCP framework coordinates the interaction between the user, the LLM, and the robotic system to ensure the generation of safe and precise task plans. These plans adhere to spatial relationships, action sequences, safety constraints, and performance metrics, while dynamically resolving ambiguities in natural language instructions.

\subsubsection{LLM Integration in the ERCP Workflow}
The ERCP leverages the LLM for the following key operations:
\begin{enumerate}
	\item \textbf{Analyzing Instructions:} The LLM identifies ambiguities, missing details, or inconsistencies in the user's instruction (e.g., "Bring me a drink").
	\item \textbf{Generating Clarification Prompts:} The LLM generates targeted questions to refine the ambiguous instruction (e.g., "What kind of drink would you like?" or "Where should I fetch the drink from?").
	\item \textbf{Iterative Refinement:} Through a back-and-forth dialogue with the user, the LLM iteratively refines the instruction until it is clear, complete, and includes all necessary spatial and contextual information (e.g., "Fetch coffee from the coffee machine and deliver it to the dining area table").
	\item \textbf{Task Plan Generation:} Once the instruction is unambiguous, the LLM translates it into a detailed task plan. This plan specifies the sequence of actions, spatial relationships, safety checks, and performance metrics for the robot to execute the task safely and efficiently.
\end{enumerate}

\subsubsection{Framework Definition}
Let $\Omega = (\mathcal{S}, \mathcal{P}, \mathcal{R}, \mathcal{L}, \mathcal{T})$ represent the ERCP framework where:
\begin{itemize}
	\item $\mathcal{S}$ is the state space, 
	\item $\mathcal{P}$ the prompt space,
	\item $\mathcal{R}$ the response space,
	\item $\mathcal{L}$ the language model function space,
	\item $\mathcal{T}$ the state transition functions.
\end{itemize}

The state $S_t \in \mathcal{S}$ at time $t$ is defined as:
\begin{equation}
	S_t = (K_t, C_t, I_t, E_t),
\end{equation}
where:
\begin{itemize}
	\item $K_t \in \mathcal{K}$ is the knowledge state,
	\item $C_t \in \mathcal{C}$ represents constraints,
	\item $I_t \in \mathcal{I}$ defines the intent,
	\item $E_t \in \mathcal{E}$ encodes the environmental context.
\end{itemize}
\newpage
\subsubsection{Core Process}
The iterative refinement process combines the structured guidance of ERCP with the generative capabilities of the LLM to achieve clear and executable instructions. The process consists of three stages:

\begin{enumerate}
	\item \textbf{Response Generation:} The LLM generates a response to the current prompt and state.
	\begin{equation}
		R_t = L(P_t, S_{t-1}, \theta) : \mathcal{P} \times \mathcal{S} \times \Theta \rightarrow \mathcal{R},
	\end{equation}
	where $\theta \in \Theta$ represents the LLM parameters.
	
	\item \textbf{State Update:} The ERCP updates the state based on the LLM-generated response.
	\begin{equation}
		S_t = f(S_{t-1}, R_t, \xi) : \mathcal{S} \times \mathcal{R} \times \Xi \rightarrow \mathcal{S},
	\end{equation}
	where $\xi \in \Xi$ encodes external context.
	
	\item \textbf{Next Prompt Generation:} The ERCP generates the next prompt for the LLM based on the updated state.
	\begin{equation}
		P_{t+1} = g(S_t, \eta) : \mathcal{S} \times H \rightarrow \mathcal{P},
	\end{equation}
	where $\eta \in H$ defines prompt generation parameters.
\end{enumerate}

The process iterates until the instruction is clear and complete. Once this is achieved, the LLM generates a detailed task plan.

\subsubsection{Example Workflow}
\textbf{Ambiguous Instruction:} "Get me a drink."  
\begin{itemize}
	\item \textbf{Step 1: Analyze Instruction:}  
	- LLM detects ambiguity: Missing parameters (e.g., type of drink, temperature, location).  
	- Generated Prompt: "What kind of drink would you like?"
	\item \textbf{Step 2: User Response:}  
	- User replies: "Coffee."
	\item \textbf{Step 3: Clarify Spatial and Handling Details:}  
	- LLM generates another prompt: "Where should I fetch the coffee from, and where should I deliver it?"  
	- User responds: "Fetch it from the coffee machine and deliver it to the dining area table."
	\item \textbf{Step 4: Generate Task Plan:}  
	- Final instruction: "Fetch coffee from the coffee machine and deliver it to the dining area table."  
\end{itemize}
\begin{figure}[h]
	\centering
	\includegraphics[scale=0.30]{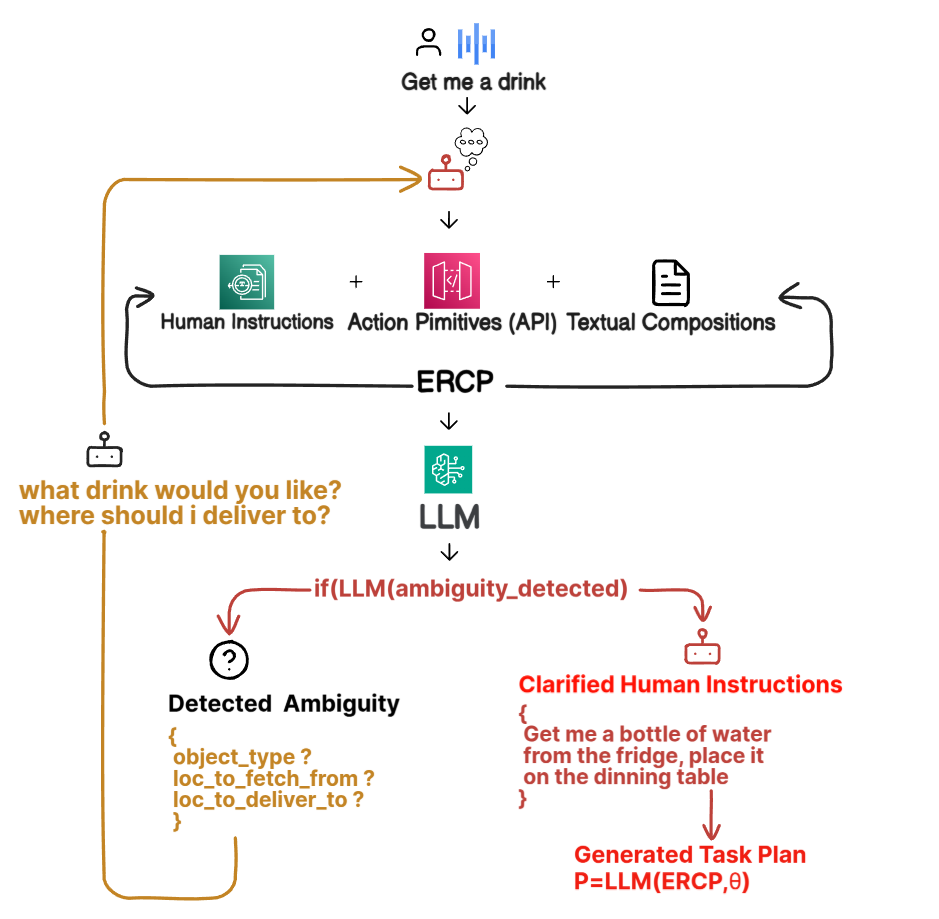}
	\caption{ERCP workflow starts with a user command, followed by the LLM analyzing for ambiguities. If ambiguities exist, the LLM generates clarification prompts, leading to user responses. This loop continues until the instruction is clear, culminating in a detailed task plan executed by the robot.}
	\label{fig:fig:enter-label}
\end{figure}

\paragraph{LLM generates the task plan focusing on}
\begin{enumerate}
	\item \textbf{Interact with Appliance:}
	\begin{itemize}
		\item Open the coffee machine carefully (force limit: $0N$).
		\item Grasp the coffee using a precision grip:
		\begin{itemize}
			\item Grip force: $7.67N$.
			\item Orientation: Upright.
			\item Safety checks: Monitor temperature ($<85^\circ C$) and ensure stability.
		\end{itemize}
		\item Close the coffee machine.
	\end{itemize}
	
	\item \textbf{Navigate to Target Location:}
	\begin{itemize}
		\item Move to the dining area table while keeping the coffee upright and monitoring temperature.
		\item Avoid obstacles with a minimum distance of $0.7m$.
		\item Maintain a maximum velocity of $0.8m/s$ and acceleration of $0.5m/s^2$.
	\end{itemize}
	
	\item \textbf{Place Item:}
	\begin{itemize}
		\item Carefully place the coffee on the dining area table.
		\item Perform surface detection to verify stability.
		\item Ensure placement force does not exceed $2N$ and the item is centered on the table.
	\end{itemize}
	
	\item \textbf{Performance Metrics:}
	\begin{itemize}
		\item Success criteria:
		\begin{itemize}
			\item Safe interaction with the coffee machine.
			\item Secure grasp of the coffee.
			\item Obstacle-free navigation.
			\item Precise and stable placement of the coffee.
		\end{itemize}
		\item Estimated time: $120s$.
	\end{itemize}
\end{enumerate}

\subsubsection{Handling Complex Ambiguities}
The ERCP framework is designed to handle more complex ambiguities through multi-step refinement. For example:
\begin{itemize}
	\item \textbf{Ambiguous Instruction:} "Clean up the room."
	\item \textbf{Step 1: Analyze Instruction:}  
	- LLM detects ambiguity: Missing parameters (e.g., which room, what objects to clean, cleaning method).  
	- Generated Prompt: "Which room should I clean, and what objects need attention?"
	\item \textbf{Step 2: User Response:}  
	- User replies: "Clean the living room, and pick up the toys on the floor."
	\item \textbf{Step 3: Clarify Spatial and Handling Details:}  
	- LLM generates another prompt: "Where should I place the toys after picking them up?"  
	- User responds: "Place them in the toy box near the couch."
	\item \textbf{Step 4: Generate Task Plan:}  
	- Final instruction: "Pick up the toys from the living room floor and place them in the toy box near the couch."  
\end{itemize}

\subsection{Embodied Knowledge Graph (EKG)}
The \textbf{Embodied Knowledge Graph (EKG)} is a structured representation of the robot's environment, implemented using Neo4j's graph database architecture \cite{neo4j2023}. It encapsulates spatial, physical, temporal, and task-specific parameters, enabling the robot to reason about its environment and validate task plans against real-time data. The EKG is formally defined as:

\[
\text{EKG} = \{(v_i, r_j, v_k) \mid v_i, v_k \in \mathcal{V}, r_j \in \mathcal{R}\},
\]

where:
\begin{itemize}
	\item \( v_i, v_k \): Vertices representing entities (e.g., objects, locations, tasks),
	\item \( r_j \): Relationships between entities (e.g., "located on," "part of").
\end{itemize}

\begin{figure}[h]
	\centering
	\includegraphics[scale=0.25]{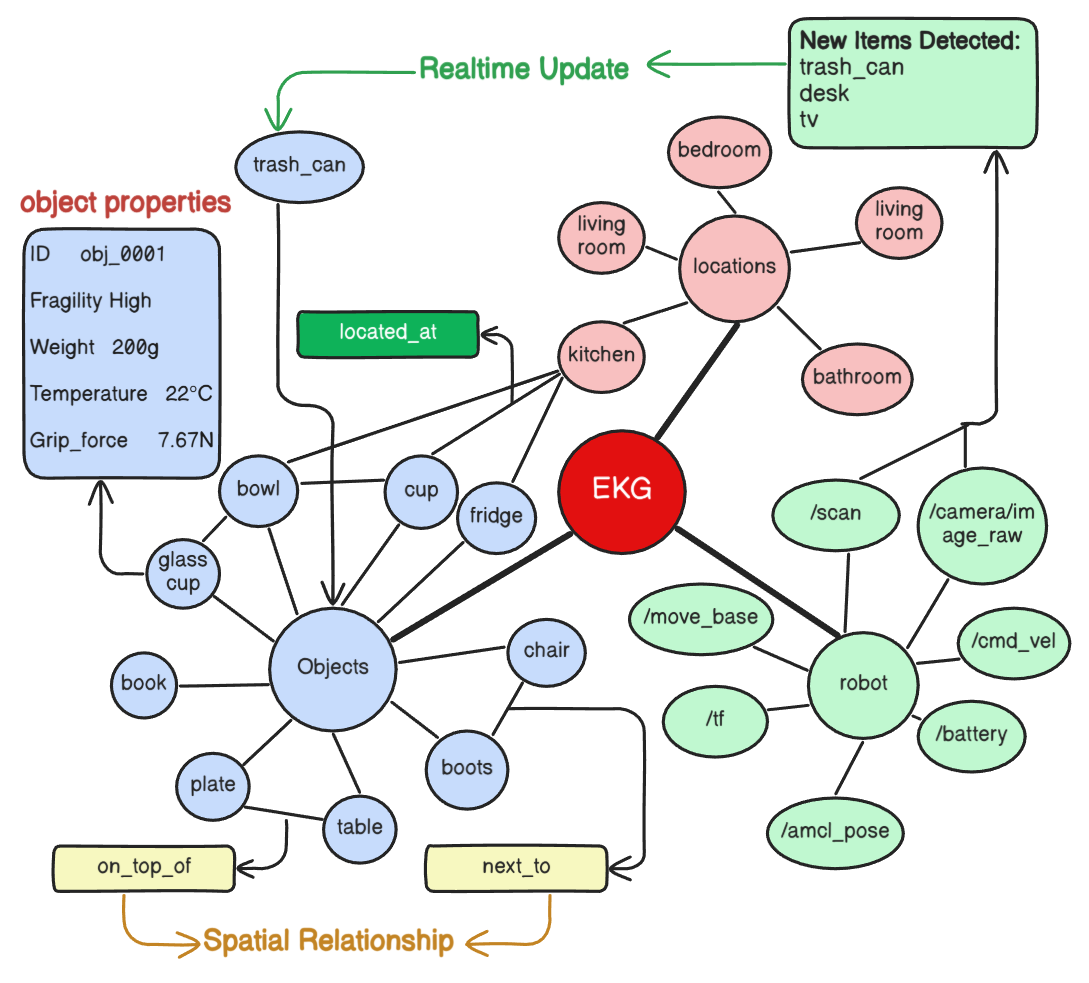}
	\caption{\textbf{Embodied Knowledge Graph (EKG):} A spatial and relational map of an indoor environment, showing entities (e.g., locations, objects, robot), their relationships (e.g., \texttt{on\_top\_of}, \texttt{next\_to}), and real-time attributes (e.g., robot position, battery status). The EKG integrates sensor data (e.g., \texttt{/scan}, \texttt{/camera/mage\_raw}) to dynamically represent the environment, enabling safe and efficient task execution.}
	\label{fig:fig:enter-label}
\end{figure}

\subsubsection{Real-Time Updates for EKG}
The EKG integrates both synchronous and asynchronous updates to balance safety and efficiency. Critical updates, such as detecting dynamic obstacles or safety violations, are handled synchronously to ensure immediate response. Less critical updates, like changes in object properties (e.g., ambient temperature), are processed asynchronously to avoid unnecessary delays.

\textbf{Synchronization Mechanism}:
\begin{itemize}
	\item \textbf{Critical Updates (Synchronous)}: For events such as newly detected obstacles or violations of safety thresholds, the EKG halts task execution and updates immediately to ensure safe operation.
	\item \textbf{Non-Critical Updates (Asynchronous)}: Updates related to minor environmental changes, such as lighting variations or ambient temperature, are processed independently in parallel with ongoing tasks.
\end{itemize}

The update strategy classifies incoming data based on priority, leveraging sensor flags and predefined thresholds to determine the appropriate update mode.

\subsubsection{Sensory Inputs and ROS Topics}
The robot's onboard sensors (e.g., cameras, LiDAR) and wall-mounted cameras provide real-time data, which is published to ROS topics. The EKG subscribes to these topics to monitor the robot's condition, position within the map, and environmental changes. The sensory inputs include:
\begin{itemize}
	\item \textbf{Robot State}: Position \((x, y, z)\), orientation \((\theta)\), velocity \((v)\), and joint configurations \((q)\).
	\item \textbf{Object Detection}: Object positions, attributes (e.g., fragility, temperature), and spatial relationships (e.g., "on top of," "next to").
	\item \textbf{Environmental Changes}: Dynamic obstacles, lighting conditions, and object state changes (e.g., movement, spills).
\end{itemize}

The sensory data is represented as a set of observations \( \mathcal{O} = \{o_1, o_2, \dots, o_n\} \), where each observation \( o_i \) corresponds to a specific sensor reading or environmental update.

\subsubsection{Object Detection and Localization}
The system employs YOLOv7 \cite{wang2023yolov7} models fine-tuned on the \textbf{RoboCup@Home dataset} \cite{wisspeintner2009robocup} and custom datasets for object detection and localization. Robot-mounted and wall-mounted cameras provide complementary viewpoints, with wall-mounted units extending coverage to robot-occluded areas. For specialized tasks like liquid pouring, custom datasets enable precise state estimation (e.g., fluid levels) for spillage prevention. Detected objects and their spatial relationships are encoded as nodes and edges in the EKG.

The EKG is updated by integrating sensory data into the graph structure. Let \( \text{EKG}_{\text{old}} = (\mathcal{V}_{\text{old}}, \mathcal{E}_{\text{old}}, \mathcal{R}_{\text{old}}, \mathcal{A}_{\text{old}}) \) represent the current state of the EKG, where:
\begin{itemize}
	\item \( \mathcal{V}_{\text{old}} \): Set of vertices (entities),
	\item \( \mathcal{E}_{\text{old}} \): Set of edges (relationships),
	\item \( \mathcal{R}_{\text{old}} \): Set of relation types,
	\item \( \mathcal{A}_{\text{old}} \): Set of attributes.
\end{itemize}

Given a set of observations \( \mathcal{O} \), the EKG is updated as follows:
\[
\text{EKG}_{\text{new}} = \text{EKG}_{\text{old}} \cup \Delta(\mathcal{O}),
\]
where \( \Delta(\mathcal{O}) \) represents the changes induced by the observations. The update process includes:
\begin{itemize}
	\item Adding new entities (e.g., detected objects):
	\[
	\mathcal{V}_{\text{new}} = \mathcal{V}_{\text{old}} \cup \{v_{\text{new}}\},
	\]
	where \( v_{\text{new}} \) represents a newly detected object.
	\item Updating attributes (e.g., object positions, temperatures):
	\[
	\mathcal{A}_{\text{new}} = \mathcal{A}_{\text{old}} \cup \{(v_i, a_j, d_j)\},
	\]
	where \( v_i \) is an entity, \( a_j \) is an attribute, and \( d_j \) is the updated value.
	\item Adding or modifying relationships (e.g., spatial relationships):
	\[
	\mathcal{E}_{\text{new}} = \mathcal{E}_{\text{old}} \cup \{(v_i, r_j, v_k)\},
	\]
	where \( v_i \) and \( v_k \) are entities, and \( r_j \) is a relationship type.
\end{itemize}

\subsubsection{Graph Attention Networks (GATs) for Prioritization}
To prioritize critical nodes and edges during task plan validation, we employ \textbf{Graph Attention Networks (GATs)}. GATs compute attention weights \(\alpha_{ij}\) for nodes \(i\) and \(j\) in the Embodied Knowledge Graph (EKG), enabling the robot to focus on the most relevant aspects of the environment. The attention weights are computed as:

\[
\alpha_{ij} = \frac{\exp\left(\text{LeakyReLU}\left(\mathbf{a}^{T}[\mathbf{W}h_{i} \| \mathbf{W}h_{j}]\right)\right)}{\sum_{k \in \mathcal{N}(i)} \exp\left(\text{LeakyReLU}\left(\mathbf{a}^{T}[\mathbf{W}h_{i} \| \mathbf{W}h_{k}]\right)\right)},
\]

where:
\begin{itemize}
	\item \(\mathbf{W} \in \mathbb{R}^{d \times d}\): Weight matrix for feature transformation,
	\item \(\mathbf{a} \in \mathbb{R}^{d}\): Learnable attention vector,
	\item \(h_{i}, h_{j} \in \mathbb{R}^{d}\): Feature representations of nodes \(i\) and \(j\),
	\item \(\mathcal{N}(i)\): Set of neighbors of node \(i\).
\end{itemize}

The feature vector \(h_i\) includes physical properties (e.g., fragility, weight) and relational properties (e.g., proximity to the robot). Safety-critical nodes are identified based on thresholds (e.g., force limits, object importance) derived empirically or via pre-trained weights.

\begin{figure}[h]
	\centering
	\includegraphics[scale=0.27]{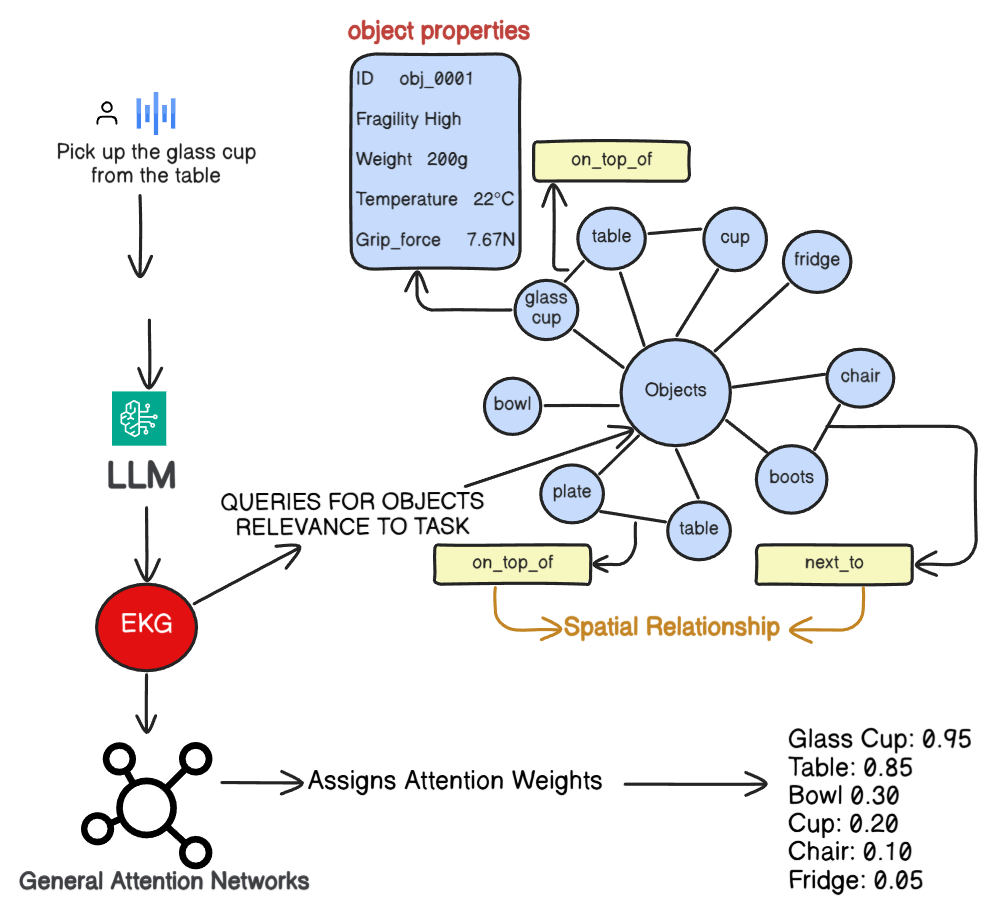}
	\caption{\textbf{General Attention Network (GAT) for Task \texttt{Pick up the glass cup}}}
	\label{fig:fig:enter-label}
\end{figure}
\FloatBarrier

\begin{figure}[h]
	\centering
	\includegraphics[scale=0.52]{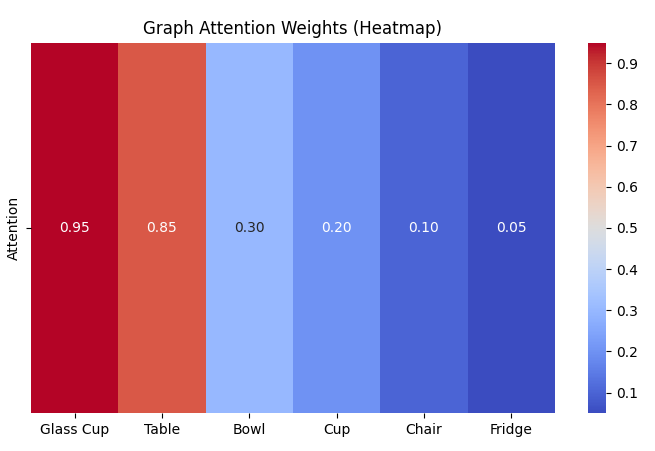}
	\caption{ The glass cup (0.95) and table (0.85) receive the highest weights due to their direct relevance and spatial relationship. Nearby objects (e.g., bowl, cup) and obstacles (e.g., chair) receive lower weights, while irrelevant objects (e.g., fridge) are assigned minimal weights, ensuring the robot focuses on safety-critical constraints.}
	\label{fig:fig:enter-label}
\end{figure}
\FloatBarrier

\subsubsection{Hamiltonian Paths for Task Sequencing}
To ensure task plans visit all critical nodes in the correct sequence, we introduce \textbf{Hamiltonian Paths}. Given a set of critical nodes \(\{v_{1}, v_{2}, \ldots, v_{n}\}\) in the Embodied Knowledge Graph (EKG), the Hamiltonian Path \(P\) is defined as:

\[
P = \arg\min_{\text{path}} \sum_{i=1}^{n-1} C(v_{i}, v_{i+1}),
\]

where \(C(v_{i}, v_{i+1})\) is the cost of traversing from node \(v_{i}\) to \(v_{i+1}\). The cost function \(C(v_{i}, v_{i+1})\) incorporates factors such as distance, safety constraints, and environmental dynamics. Specifically, it is defined as:

\begin{align*}
	C(v_{i}, v_{i+1}) &= w_1 \cdot \text{distance}(v_{i}, v_{i+1}) \\
	&\quad + w_2 \cdot \text{safety\_constraints}(v_{i}, v_{i+1}) \\
	&\quad + w_3 \cdot \text{environmental\_dynamics}(v_{i}, v_{i+1}),
\end{align*}

where \(w_1\), \(w_2\), and \(w_3\) are weights empirically derived through sensitivity analysis to balance these factors.

\begin{figure}[h]
	\centering
	\includegraphics[scale=0.29]{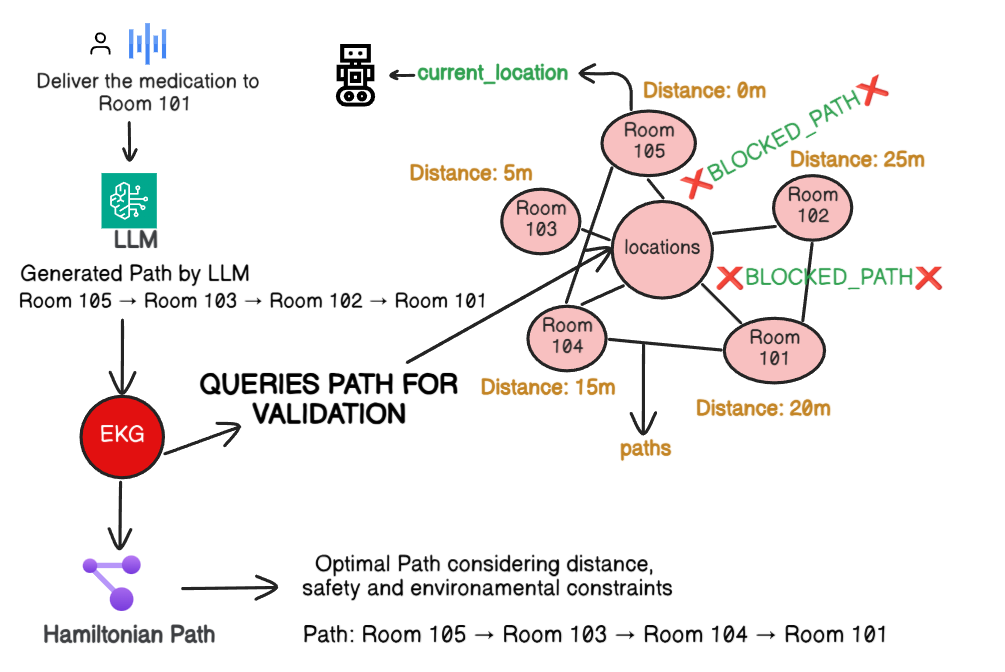}
	\caption{\textbf{Hamiltonian Path for Medication Delivery:} The diagram illustrates the optimal path for delivering medication, following the sequence \texttt{Room 105 $\rightarrow$ Room 103 $\rightarrow$ Room 102 $\rightarrow$ Room 101}. The Hamiltonian Path ensures the robot visits all critical locations in the correct order while minimizing distance (5m) and adhering to safety constraints.}
	\label{fig:fig:enter-label}
\end{figure}
\FloatBarrier

\subsubsection{Conflict Resolution Between Commands and Environmental Constraints}
The EKG resolves conflicts between commands and environmental constraints using graph-based reasoning and prioritization. When a command is issued, the EKG validates it against the environment's current state, represented as nodes and edges. Conflicts occur when commands violate safety thresholds, spatial constraints, or dynamic conditions. The EKG resolves conflicts through the following steps:

\begin{enumerate}
	\item \textbf{Conflict Detection:} 
	The EKG identifies conflicts by comparing command requirements with the environment. For example, if a navigation command involves a blocked pathway, the EKG detects the conflict by querying the graph for obstacles.
	
	\item \textbf{Prioritization Using GATs:} 
	Graph Attention Networks (GATs) prioritize conflicting constraints. Safety-critical nodes (e.g., obstacles, force limits) receive higher attention weights, ensuring precedence over less critical constraints. For instance, if moving an object exceeds a force limit, the EKG flags the command as invalid.
	
	\item \textbf{Alternative Path Generation:} 
	When conflicts arise, the EKG generates alternative paths or actions. For example, if a dynamic obstacle blocks navigation, the EKG computes a new Hamiltonian Path to avoid the obstacle while minimizing traversal cost, incorporating distance, safety, and environmental dynamics.
	
	\item \textbf{User Feedback and Refinement:} 
	If conflicts cannot be resolved autonomously, the EKG uses the Embodied Robotic Control Prompt (ERCP) framework to request user input. For example, if an object is out of reach, the user is prompted to specify an alternative location or adjust task parameters.
\end{enumerate}

\paragraph{Example Scenario: Conflict Resolution}
Consider a command: "Fetch the coffee cup from the table and deliver it to the kitchen counter." If the table is obstructed by a moving person, the EKG resolves the conflict as follows:
\begin{itemize}
	\item \textbf{Conflict Detection:} The EKG identifies the obstacle and flags the path as blocked.
	\item \textbf{Prioritization:} The obstacle node is prioritized using GATs.
	\item \textbf{Alternative Path Generation:} The EKG computes a new Hamiltonian Path to avoid the obstacle.
	\item \textbf{User Feedback:} If no feasible path exists, the user is prompted to wait or specify an alternative location.
\end{itemize}

By integrating conflict detection, prioritization, and alternative path generation, the EKG ensures safe and efficient command execution in dynamic environments.

\subsection{Task Plan Generation}

Given the human instruction \((c)\): ``pick up the mug from the dining table, place it on the serving tray'', the system constructs the Embodied Robotic Control Prompt (ERCP) using a default prompt template. The ERCP serves as a structured input for the Large Language Model (LLM), which generates a detailed task plan. The task plan generated by the LLM can be broken down into three main components: \textbf{spatial relationships}, \textbf{task plan (actions and their sequences)}, and \textbf{performance metrics and safety criteria}.

\subsubsection{Spatial Relationships}

The spatial relationships define the object's location, orientation, and distances between the source and destination. For the given instruction, the LLM generates the following spatial relationships:

\begin{verbatim}
	{"spatial_relationships": {
			"object": "mug",
			"source": {
				"location": "dining table",
				"container_or_surface": null
			},
			"destination": {
				"location": "serving tray",
				"container_or_surface": null
			},
			"distance_to_source": "2.88m",
			"distance_to_destination": "1.19m",
			"orientation_at_source": "upright",
			"orientation_at_destination": "upright",
			"environmental_factors": {
				"lighting_conditions": "moderate",
				"surface_texture": "wooden",
				"ambient_temperature": "22°C"
			}}
\end{verbatim}

These relationships ensure the system understands the object's current and target positions, as well as the spatial constraints for navigation and placement.

\subsubsection{Task Plan (Actions and Their Sequences)}

The task plan consists of a sequence of actions required to complete the instruction. Each action is associated with specific parameters and safety checks. The LLM generates the following task plan with four key action sequences \texttt{perception}, \texttt{grasping the object}, \texttt{navigation}, and \texttt{placement}. Each action is designed with specific parameters, including safety measures to ensure the robot's performance.\\

\textbf{Description of Perception:} The \texttt{perception} action involves detecting the object (in this case, a mug), locating it on a surface (such as a dining table), ensuring the object is accessible, and scanning the surrounding area for potential obstacles. It also includes safety checks to monitor the workspace, detect obstacles, and avoid collisions.
\begin{verbatim}
		"actions": [
		{
			"action": "perception",
			"parameters": {
				"subtasks": [
				"detect_object(mug)", 
				"locate_surface_or_container(dining table)", 
				"verify_object_accessibility", 
				"scan_surrounding_area"
				],
				"safety_checks": {
					"workspace_monitoring": "active", 
					"obstacle_detection": "enabled", 
					"collision_detection": "active"
				}}}],
\end{verbatim}

\textbf{Description of Grasping the Object:} The \texttt{grasp\_object} action is responsible for grasping the mug with a defined grip type (power grip), applying a specified force, and adjusting the approach angle. The safety parameters ensure that force remains within limits, monitoring the force continuously while detecting slip and monitoring finger pressure.

\begin{verbatim}
	{"action": "grasp_object",
		"parameters": {
			"object_type": "mug", 
			"grip_type": "power", 
			"grip_parameters": {
				"force": "15.15N", 
				"approach_angle": "78deg", 
				"contact_points": [
				"finger1", 
				"finger2"
				]},
			"safety_parameters": {
				"max_force": "37.5N", 
				"force_monitoring": "continuous", 
				"slip_detection": "enabled", 
				"finger_pressure_sensors": "enabled"
			}}},
\end{verbatim}

\textbf{Description of Navigation:} The \texttt{navigate\_to\_destination} action directs the robot to the target location (the serving tray). The robot must maintain a safe distance from obstacles and adapt its speed based on environmental conditions, ensuring smooth and safe navigation while holding the object.

\begin{verbatim}
	{
		"action": "navigate_to_destination",
		"parameters": {
			"target_location": "serving tray", 
			"nav_type": "global_path", 
			"holding_object": true, 
			"safety_constraints": {
				"min_obstacle_distance": "0.7m", 
				"max_velocity": "0.8m/s", 
				"required_clearance": "0.8m", 
				"adaptive_velocity_control": "enabled"
	}}},
\end{verbatim}

\textbf{Description of Placement:} The \texttt{place\_object} action ensures the object is placed precisely on a surface with specific parameters, such as the final height, orientation, and angle of placement. Safety checks for maximum force, surface detection, and stability are active to ensure the object is placed securely.

\begin{verbatim}
	{
		"action": "place_object",
		"parameters": {
			"target_surface": null, 
			"placement_type": "precise", 
			"placement_parameters": {
				"approach_height": "0.01m", 
				"final_height": "0.00m", 
				"orientation": "upright", 
				"angle_of_placement": "0deg"
			},
			"safety_parameters": {
				"max_place_force": "2.0N", 
				"surface_detection": "active", 
				"stability_check": "enabled", 
				"surface_pressure_monitoring": "enabled"
			}
	}}]},
\end{verbatim}
	
\textbf{Performance Metrics and Safety Criteria:}

The performance and safety criteria are generated based on the task plan and in accordance with the human instruction, helping to determine the safety criteria that must be evaluated for the success of the task.

\textbf{Success Criteria:} This section includes the key indicators of task success, such as safe navigation, secure grasping, and correct placement.

\begin{lstlisting}[basicstyle=\ttfamily\small, breaklines=true]
	{
		"success_criteria": "navigation completed safely, object grasped securely, object transported safely, object placed correctly, task completed within time"
	}
\end{lstlisting}

\textbf{Safety Metrics:} These metrics ensure that the robot's performance stays within predefined safety parameters, such as collision avoidance and force limits.

\begin{lstlisting}[basicstyle=\ttfamily\small, breaklines=true]
	{
		"safety_metrics": "collision avoidance: verified, force limits maintained: verified, workspace monitoring: active, object stability: verified"
	}
\end{lstlisting}

\textbf{Estimated Metrics:} These include metrics like the total distance traveled, estimated time for task completion, and energy consumption.

\begin{lstlisting}[basicstyle=\ttfamily\small, breaklines=true]
	{
		"estimated_metrics": "total distance: 8.3m, estimated time: 61s, energy consumption: 21.6Wh"
	}
\end{lstlisting}

\textbf{Required Monitoring:} This section defines the aspects of the system that must be continuously monitored to ensure safe and efficient task execution.

\begin{lstlisting}[basicstyle=\ttfamily\small, breaklines=true]
	{
		"required_monitoring": "object tracking, force feedback, obstacle detection, battery level"
	}
\end{lstlisting}
	
The task plan generation process can be formally defined as:
\[
\text{ERCP} = f(c, \mathcal{A}, \mathcal{T}),
\]
where:
\begin{itemize}
	\item \( c \) is the user command,
	\item \( \mathcal{A} \) is the set of action primitives (APIs),
	\item \( \mathcal{T} \) is the set of textual clarifications and context.
\end{itemize}

The Embodied Robotic Control Prompt (ERCP) is then used as input to the Large Language Model (LLM) to generate the task plan:
\[
\mathcal{P} = \text{LLM}(\text{ERCP}, \theta),
\]
where:
\begin{itemize}
	\item \( \mathcal{P} \) is the generated task plan,
	\item \( \theta \) represents the parameters of the LLM.
\end{itemize}

The task plan \( \mathcal{P} \) is a sequence of actions \( \{a_1, a_2, \dots, a_n\} \), where each action \( a_i \) is associated with specific parameters and constraints derived from the ERCP and validated against the Embodied Knowledge Graph (EKG).

\subsection{Task Plan Validation}
The raw task plan generated by the LLM lacks consideration for environmental state changes, object attribute alterations, or spatial relationships. The Embodied Knowledge Graph (EKG) validates the plan by extracting place data, querying the EKG to form a subgraph, and analyzing it using Graph Attention Networks (GATs). 

\begin{figure}[H]
	\raggedright 
	\includegraphics[scale=0.35]{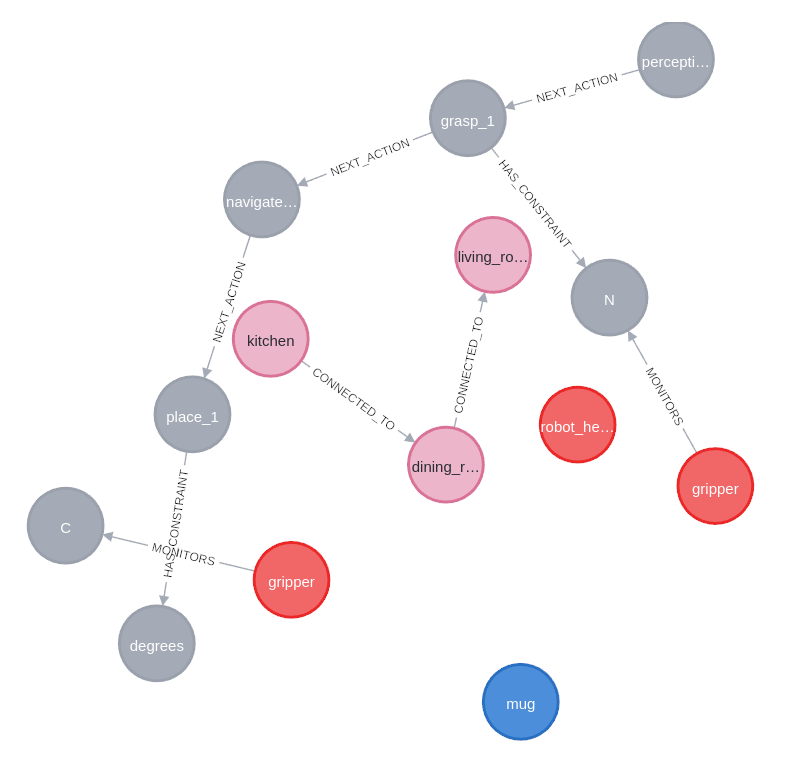}
	\caption{A subgraph of the EKG queried following the task plan generated by the LLM}
	\label{fig:fig:enter-label}
\end{figure}
\FloatBarrier

GATs compute attention weights for nodes and edges, prioritizing critical environmental aspects like object fragility, obstacle presence, and spatial relationships. Specifically, GATs leverage self-attention mechanisms to assign higher weights to nodes and edges that are most relevant to the task at hand. For example, in a task involving fragile objects, GATs prioritize nodes representing fragility constraints and edges representing spatial relationships that ensure safe manipulation. This attention mechanism allows the system to focus on the most critical aspects of the environment, ensuring that safety and efficiency constraints are met.

\subsubsection{Querying Valid Action Sequences}
The system identifies valid action sequences by querying the EKG for paths between pending actions, ensuring adherence to constraints and dependencies:

\begin{verbatim}
	action_sequence_query = """
	MATCH path = (start:Action)
	-[:NEXT_ACTION*]->(end:Action)
	WHERE start.status = 'pending'
	RETURN path;
	"""
\end{verbatim}

\textbf{Description:} Retrieves pending action sequences (e.g., \texttt{perception\_1} $\rightarrow$ \texttt{grasp\_1} $\rightarrow$ \texttt{navigate\_1} $\rightarrow$ \texttt{place\_1}), ensuring logical and executable order.

\subsubsection{Validating Constraints for Actions}
Constraints for actions (e.g., force limits, safety thresholds) are validated by querying the EKG for sensor data and constraint values:

\begin{verbatim}
	constraint_validation_query = """
	MATCH (a:Action)-[:HAS_CONSTRAINT]
	->(c:Constraint)<-[:MONITORS]-(s:Sensor)
	WHERE a.id = 'grasp_1'
	RETURN a.id, c.type, c.max_value, s.status;
	"""
\end{verbatim}

\textbf{Description:} Retrieves constraints for \texttt{grasp\_1} (e.g., max force: 37.5N) and sensor status, ensuring adherence to safety limits.

\subsubsection{Finding Paths Between Locations}
Optimal paths for navigation tasks are identified by querying the EKG for spatial relationships and environmental conditions:

\begin{verbatim}
	location_path_query = """
	MATCH path = shortestPath((start:Location)-
	[:CONNECTED_TO*1..10]->(end:Location))
	WHERE start.name = 'kitchen' 
	AND end.name = 'living_room'
	RETURN path;
	"""
\end{verbatim}

\textbf{Description:} Retrieves the shortest path (e.g., kitchen $\rightarrow$ dining room $\rightarrow$ living room), considering factors like floor type and lighting.

\subsubsection{Checking Object Properties and Constraints}
Object-specific properties and constraints are validated to ensure task feasibility and safety:

\begin{verbatim}
	object_properties_query = """
	MATCH (o:Object)
	WHERE o.name = 'mug'
	RETURN o.name, o.fragility, o.type, 
	o.max_value, o.material, 
	o.orientation, o.state;
	"""
\end{verbatim}

\textbf{Description:} Retrieves properties for the \texttt{mug} (e.g., fragility: 0.7, material: ceramic, orientation: upright, state: empty).

\subsection{Validation Mechanisms}
The validation of task plans involves checking whether all actions satisfy the safety constraints encoded in the EKG. The validation function \(V\) is formally defined as:

\[
V(p_i, S, S') = 
\begin{cases}  
	1 & \text{if } p_i \text{ satisfies all constraints in state } S', \\  
	0 & \text{otherwise},
\end{cases}
\]

where:
\begin{itemize}
	\item \(p_i\): A placeholder in the task plan,
	\item \(S\): Initial environmental state,
	\item \(S'\): Environmental state incorporating real-time sensor updates.
\end{itemize}

The task plan is executed only if \(V(p_i, S, S') = 1\) for all placeholders. If verification fails during the initial validation, the task is not executed. During task execution, if a safety violation is detected, operations are halted immediately, and the task is re-evaluated. Execution resumes only if the re-evaluation passes all safety criteria checks. The validation process includes the following safety constraints:

\subsubsection{Force and Grasping Constraints}
The force applied during grasping or manipulation must not exceed the object's fragility limits or the robot's maximum force capabilities. The safety condition is:

\[
F_{\text{applied}} \leq F_{\text{max}}.
\]

The validation function \(V_{\text{force}}(a)\) is:

\[
V_{\text{force}}(a) = \mathbb{I}(F_{\text{applied}} \leq F_{\text{max}}).
\]

\subsubsection{Temperature Constraints}
The temperature of objects being handled must remain within safe limits. The safety condition is:

\[
T_{\text{object}} \leq T_{\text{safe}}.
\]

The validation function \(V_{\text{temperature}}(a)\) is:

\[
V_{\text{temperature}}(a) = \mathbb{I}(T_{\text{object}} \leq T_{\text{safe}}).
\]

\subsubsection{Collision Avoidance}
The robot must avoid collisions with obstacles, humans, or other objects. The safety condition is:

\[
d_{\text{obstacle}} \geq d_{\text{min}}.
\]

The validation function \(V_{\text{collision}}(a)\) is:

\[
V_{\text{collision}}(a) = \mathbb{I}(d_{\text{obstacle}} \geq d_{\text{min}}).
\]

\subsubsection{Stability and Orientation Constraints}
Objects being manipulated must remain stable and properly oriented to prevent spills, drops, or damage. Let \(\theta_{\text{object}}\) be the current tilt angle of the object, and \(\theta_{\text{max}}\) be the maximum allowable tilt angle. The safety condition is:

\[
\theta_{\text{object}} \leq \theta_{\text{max}}.
\]

The validation function \(V_{\text{stability}}(a)\) is:

\[
V_{\text{stability}}(a) = \mathbb{I}(\theta_{\text{object}} \leq \theta_{\text{max}}).
\]

\subsubsection{Temporal Constraints}
The robot must complete tasks within a specified time frame to ensure efficiency and prevent delays. Let \(t_{\text{elapsed}}\) be the time elapsed since the task began, and \(t_{\text{max}}\) be the maximum allowable time. The safety condition is:

\[
t_{\text{elapsed}} \leq t_{\text{max}}.
\]

The validation function \(V_{\text{time}}(a)\) is:

\[
V_{\text{time}}(a) = \mathbb{I}(t_{\text{elapsed}} \leq t_{\text{max}}).
\]

\subsection{Problem Formulation}
The primary goal of this framework is to enable service robots to operate safely and efficiently in dynamic and unseen environments. To achieve this, the robot's task planning and execution are formalized as a sequence of discrete actions \( A = \{a_1, a_2, \dots, a_n\} \), where each action \( a_i \) contributes to the successful completion of the task while adhering to safety constraints.

The robot operates within an environment \( E \), which includes an array of objects \( O = \{o_1, o_2, \dots, o_m\} \). Each object \( o_i \) is characterized by spatial coordinates \( (x_i, y_i) \), physical attributes \( P = \{p_1, p_2, \dots, p_k\} \), and dynamic properties such as temperature and stability. The robot employs a set of sensors \( \text{Sens} = \{s_1, s_2, \dots, s_l\} \), including Intel RealSense cameras, to provide real-time sensory feedback about the environment. This feedback is crucial for maintaining safety and achieving task objectives in dynamic and unpredictable settings.

The task consists of multiple steps, such as navigating through the environment, interacting with objects, and handling temperature-sensitive items. The primary objective is to generate an optimal sequence of actions \( A^* \) that satisfies all safety constraints, preserves the integrity of objects, and ensures the successful completion of the task while maintaining efficiency.

\subsubsection{Objective Function}
The task is formalized as the minimization of a multi-dimensional objective function that accounts for spatial relationships, physical attributes, sensory inputs, safety regulations, and performance metrics. This objective function is denoted as \( f(A, E, O, P, \text{Sens}, T) \), and the optimal action sequence \( A^* \) is determined as follows:
\[
A^* = \arg\min_{A} f(A, E, O, P, \text{Sens}, T),
\]
where the objective function \( f(A, E, O, P, \text{Sens}, T) \) is a weighted sum of terms that quantify various factors, including:
\[
\begin{array}{l}
	f(A, E, O, P, \text{Sens}, T) = \\
	\alpha_1 \cdot f_{\text{path}}(A, E) + \\
	\alpha_2 \cdot f_{\text{force}}(A, O) + \\
	\alpha_3 \cdot f_{\text{temperature}}(A, O, \text{Sens}) + \\
	\alpha_4 \cdot f_{\text{stability}}(A, O, P) + \\
	\alpha_5 \cdot f_{\text{time}}(A).
\end{array}
\]
Here:
\begin{itemize}
	\item \( f_{\text{path}}(A, E) \): Cost associated with path planning and navigation, considering distance and velocity constraints. This term ensures that the robot follows a feasible and safe path through the environment.
	\item \( f_{\text{force}}(A, O) \): Ensures that force limits are maintained during object interaction and grasping. This term is validated using the EKG, which encodes object fragility and maximum force thresholds.
	\item \( f_{\text{temperature}}(A, O, \text{Sens}) \): Ensures that the temperature-sensitive properties of the item are respected. The EKG provides real-time updates on object temperatures, and the system prioritizes nodes representing temperature constraints using GATs.
	\item \( f_{\text{stability}}(A, O, P) \): Ensures that the item remains stable during manipulation and placement. Stability constraints are validated against the EKG, which encodes object stability thresholds and spatial relationships.
	\item \( f_{\text{time}}(A) \): Represents the time taken to execute the task, which must be minimized to enhance operational efficiency. This term is influenced by the Hamiltonian Path, which ensures the correct sequence of actions is followed to minimize delays.
\end{itemize}

\subsubsection{Determining the Weights}
The weights \( \alpha_i \) in the objective function are determined based on the relative importance of each factor. For example, in tasks involving fragile objects, stability constraints may be prioritized over time efficiency. The weights are computed as:
\[
\alpha_i = \frac{1}{n} \sum_{j=1}^n w_j \cdot \mathbb{I}(\text{constraint}_j),
\]
where:
\begin{itemize}
	\item \( w_j \): Importance of the \( j \)-th constraint, determined through domain knowledge, user preferences, and empirical data,
	\item \( \mathbb{I}(\text{constraint}_j) \): Indicator function that returns 1 if the \( j \)-th constraint is active and 0 otherwise.
\end{itemize}

For example, in the task of serving hot tea, the weight for temperature constraints \( w_3 \) may be set higher than the weight for time constraints \( w_5 \), ensuring that the tea is served at the correct temperature even if it takes slightly longer.

\subsubsection{Ensuring Correct Task Sequence with Hamiltonian Paths}
To ensure that the robot follows the correct sequence of actions, the system employs \textbf{Hamiltonian Paths}. Given a set of critical nodes \(\{v_{1}, v_{2}, \ldots, v_{n}\}\) in the EKG, the Hamiltonian Path \(P\) ensures that the robot visits all critical nodes in the correct order. The Hamiltonian Path is defined as:

\[
P = \arg\min_{\text{path}} \sum_{i=1}^{n-1} C(v_{i}, v_{i+1}),
\]

where \(C(v_{i}, v_{i+1})\) is the cost of transitioning from node \(v_{i}\) to \(v_{i+1}\). The cost function \(C\) incorporates factors such as safety constraints, environmental dynamics, and task dependencies. The Hamiltonian Path ensures that the robot follows the correct sequence of actions, such as navigating to the coffee machine, picking up the tea, and delivering it to the lounge table, while adhering to safety and efficiency constraints.

\subsubsection{Handling Dynamic and Unseen Environments}
The framework is designed to handle dynamic and unseen environments by continuously updating the EKG based on real-time sensory inputs. For example, if a new obstacle is detected, the EKG is updated to reflect the obstacle's position, and the task plan is revised to avoid collisions. The updated EKG is used to re-evaluate the objective function and generate a new optimal action sequence \( A^* \).

The process of handling dynamic changes can be formalized as:
\[
A^*_{\text{new}} = \arg\min_{A} f(A, E_{\text{new}}, O_{\text{new}}, P_{\text{new}}, \text{Sens}, T),
\]
where \( E_{\text{new}} \), \( O_{\text{new}} \), and \( P_{\text{new}} \) represent the updated environment, objects, and properties based on real-time sensory data.

If a new obstacle is detected during task execution, the system updates the EKG, re-evaluates the objective function, and generates a new optimal action sequence \( A^*_{\text{new}} \) that avoids the obstacle while satisfying all safety constraints.

\subsection{Task Execution and Markov Decision Process Framework}
The execution of tasks is formalized using a \textbf{Markov Decision Process (MDP)} framework, which provides a structured approach for sequential decision-making under uncertainty. This framework integrates with the \textbf{Embodied Knowledge Graph (EKG)} to enable task planning and execution in dynamic environments. The MDP framework ensures that the robot can adapt its actions based on real-time environmental changes and task requirements.

\subsubsection{MDP Formalization}
The MDP is defined as a tuple \( M = (S, A, T, R, \gamma) \), where:
\begin{itemize}
	\item \( S \) represents the state space, which includes:
	\begin{itemize}
		\item Robot joint configurations \( q \in \mathbb{R}^n \),
		\item Object poses \( p_o \in SE(3) \),
		\item Environmental parameters \( e \in \mathbb{R}^m \),
		\item Task progress indicators \( \tau \in [0, 1] \).
	\end{itemize}
	
	\item \( A \) defines the action space, consisting of:
	\begin{itemize}
		\item Primitive actions \( a_p \in A_p \) (e.g., grasp, push, lift),
		\item Composite actions \( a_c \in A_c \) derived from action sequences,
		\item Control parameters \( \theta \in \Theta \) for action execution.
	\end{itemize}
	
	\item \( T: S \times A \times S \rightarrow [0, 1] \) represents the transition function:
	\begin{itemize}
		\item \( T(s'|s, a) = P(s_{t+1} = s'|s_t = s, a_t = a) \),
		\item Incorporates environmental dynamics and action uncertainties,
		\item Updated through the EKG based on execution outcomes.
	\end{itemize}
	
	\item \( R: S \times A \times S \rightarrow \mathbb{R} \) defines the reward function:
	\[
	R(s, a, s') = w_1 r_i + w_2 r_p + w_3 r_s,
	\]
	where:
	\begin{itemize}
		\item \( r_i(s, a, s') \) represents immediate rewards for action completion,
		\item \( r_p(\tau) \) denotes progress rewards for task advancement,
		\item \( r_s(s) \) indicates safety penalties for approaching constraints.
	\end{itemize}
	
	\item \( \gamma \in [0, 1) \) is the discount factor balancing immediate and future rewards.
\end{itemize}

\subsubsection{Task Decomposition and Action Selection}
Tasks are decomposed into subtasks using a hierarchical structure:
\begin{enumerate}
	\item \textbf{High-level task representation}:
	\[
	\mathcal{T} = \{(g_1, \phi_1), \ldots, (g_n, \phi_n)\},
	\]
	where \( g_i \) represents subtask goals and \( \phi_i \) defines constraints.
	
	\item \textbf{Task plan selection policy} \( \pi: S \rightarrow A \):
	\begin{itemize}
		\item Derived through value iteration:
		\[
		V^*(s) = \max_{a \in A} \left\{ R(s, a) + \gamma \sum_{s' \in S} T(s'|s, a) V^*(s') \right\},
		\]
		\item Policy extraction:
		\[
		\pi^*(s) = \arg\max_{a \in A} \left\{ R(s, a) + \gamma \sum_{s' \in S} T(s'|s, a) V^*(s') \right\}.
		\]
	\end{itemize}
	
	\item \textbf{Action sequence validation through EKG}:
	\begin{itemize}
		\item Feasibility check: \( \mathcal{F}(\pi, \phi) \rightarrow \{0, 1\} \),
		\item Safety verification: \( \mathcal{V}(\pi, s) \rightarrow \{0, 1\} \).
	\end{itemize}
\end{enumerate}

\subsubsection{Feedback Loop and Learning}
The feedback loop integrates execution outcomes to refine both the MDP components and the EKG:
\begin{enumerate}
	\item \textbf{State-Action-Reward Collection}:
	\begin{itemize}
		\item Execution traces: \( \xi = \{(s_t, a_t, r_t, s_{t+1})\}_{t=0}^T \),
		\item Performance metrics: \( \mathcal{M}(\xi) = \{\text{success, efficiency, safety}\} \).
	\end{itemize}
	
	\item \textbf{Model Update Rules}:
	\begin{itemize}
		\item Transition function update:
		\[
		T'(s'|s, a) = (1 - \alpha) T(s'|s, a) + \alpha \mathbb{I}_{(s, a, s')}(\xi),
		\]
		\item Reward function refinement:
		\[
		R'(s, a, s') = R(s, a, s') + \beta \Delta(\mathcal{M}(\xi)),
		\]
		\item EKG knowledge update:
		\[
		\text{EKG}' = \text{EKG} \oplus \{\{\xi, \mathcal{M}(\xi)\}\}.
		\]
	\end{itemize}
	
	\item \textbf{Policy Improvement}:
	\begin{itemize}
		\item Value function update using collected experiences,
		\item Policy refinement through gradient-based optimization:
		\[
		\theta' = \theta + \eta \nabla_\theta J(\pi_\theta).
		\]
	\end{itemize}
\end{enumerate}

\subsection{Real-Time Validation with GATs}
Graph Attention Networks (GATs) address the challenge of real-time validation by dynamically prioritizing safety-critical nodes in the Embodied Knowledge Graph (EKG), such as fragile objects (e.g., glassware), dynamic obstacles (e.g., moving humans), or temperature-sensitive items (e.g., hot beverages). This prioritization is achieved through attention weights computed as:
\[
\alpha_{ij} = \frac{\exp\left(\text{LeakyReLU}\left(\mathbf{a}^T [\mathbf{W}h_i \| \mathbf{W}h_j]\right)\right)}{\sum_{k \in \mathcal{N}(i)} \exp\left(\text{LeakyReLU}\left(\mathbf{a}^T [\mathbf{W}h_i \| \mathbf{W}h_k]\right)\right)},
\]
where \(h_i, h_j\) are node features (e.g., fragility, temperature, proximity to the robot), \(\mathbf{W}\) is a learnable weight matrix, and \(\mathbf{a}\) is an attention vector. 

\noindent \textbf{Key Implementation Details:}
\begin{itemize}
	\item \textbf{Node Features}: Features are derived from sensor data (e.g., LiDAR, cameras) and predefined object properties in the EKG. For example:
	\begin{itemize}
		\item Fragility: Binary flag (1 for fragile items like vases, 0 otherwise) \cite{ortenzi2021object}.
		\item Proximity: Euclidean distance between the robot and the object, updated at 10Hz.
		\item Temperature: Real-time thermal sensor readings for items like coffee mugs \cite{hu2024deploying}.
	\end{itemize}
	
	\item \textbf{Attention Weights in Practice}: During a "serve coffee" task, GATs assign higher weights to nodes representing the coffee mug (\(\alpha = 0.92\)) and the target table (\(\alpha = 0.85\)) than to non-critical nodes like ambient lighting (\(\alpha = 0.12\)). This ensures the robot focuses on handling the mug safely while monitoring spill risks.
\end{itemize}

\noindent \textbf{Safety-Driven Validation:}
The prioritized nodes trigger real-time validation checks. For example:
\begin{itemize}
	\item If a mug’s temperature exceeds a safety threshold (e.g., \(>85^\circ\)C), the EKG updates its node properties, and GATs increase its attention weight, forcing the robot to adjust its grip force or trajectory.
	\item If a dynamic obstacle (e.g., a human) enters the robot’s path, the obstacle’s proximity feature updates, and GATs prioritize collision avoidance by rerouting the task plan within \(<0.5s\) \cite{vasic2013safety}.
\end{itemize}

\section{Experiments}

To evaluate the safety compliance and task execution effectiveness of our proposed framework, we conducted a comprehensive set of experiments across two robotic platforms. Our evaluation includes comparisons between (i) \textbf{ERCP + LLM for task plan generation} and state-of-the-art LLM-based task planners, and (ii) \textbf{EKG for real-time knowledge-based validation} against traditional knowledge-based methods. Additionally, we performed human evaluations to assess the usability and effectiveness of the framework in real-world scenarios.

\subsection{Experimental Setup}
\label{subsec:experimental_setup}

\subsubsection{Hardware and Software Specifications}
Our framework was deployed on two robotic platforms:
\begin{itemize}
	\item \textbf{Yahboom Jetson TX2}: Mobile robot with 6-DOF arm, Intel RealSense D435 camera, and ROS Noetic-based navigation stack.
	\item \textbf{Realman Intelligent Robot}: Stationary 7-DOF manipulator with parallel gripper and custom inverse kinematics solver.
\end{itemize}
Implementation details:
\begin{itemize}
	\item \textbf{Large Language Model}: Fine-tuned \texttt{meta-llama/Llama-3.2-3B-Instruct} for robotic task execution.
	\item \textbf{Knowledge Graph Engine}: Neo4j 
	\item \textbf{Perception System}: Intel RealSense D435 + 360° LiDAR for environmental sensing.
\end{itemize}

\subsubsection{Dataset and Task Categories}
Evaluation conducted using \textbf{28,715 human instructions} annotated with:
\begin{itemize}
	\item Task specifications and safety constraints.
	\item Environmental metadata (dynamic obstacles, temperature sensitivity).
\end{itemize}

Three primary task categories:
\begin{itemize}
	\item \textbf{Navigation}: ``Navigate to dining area avoiding obstacles.''
	\item \textbf{Manipulation}: ``Handle fragile cup placement on tray.''
	\item \textbf{Multi-Step}: ``Fetch coffee and serve at table 5.''
\end{itemize}

\subsection{Evaluation Metrics}
Quantitative assessment using five key metrics:
\begin{itemize}
	\item \textbf{Ambiguity Resolution Rate (ARR)}: $\frac{\text{Successfully clarified instructions}}{\text{Total ambiguous instructions}} \times 100$.
	\item \textbf{Task Success Rate (TSR)}: $\frac{\text{Completed tasks}}{\text{Attempted tasks}} \times 100$.
	\item \textbf{Safety Compliance Rate (SCR)}: $\frac{\text{Safe task executions}}{\text{Total executions}} \times 100$.
	\item \textbf{Processing Time (PT)}: $\frac{\text{Total refinement time}}{\text{Number of tasks}}$.
	\item \textbf{Error Mitigation Rate (EMR)}: $\frac{\text{Corrected unsafe attempts}}{\text{Total unsafe attempts}} \times 100$.
\end{itemize}

\subsection{Comparative Analysis: ERCP + LLM vs. Baselines}
\begin{table}[h]
	\centering
	\caption{Performance Comparison of Task Planning Methods}
	\renewcommand{\arraystretch}{1.2} 
	\resizebox{\columnwidth}{!}{
		\begin{tabular}{lccccc}
			\hline
			\textbf{Method} & \textbf{ARR (\%)} & \textbf{TSR (\%)} & \textbf{SCR (\%)} & \textbf{PT (s)} & \textbf{EMR (\%)} \\
			\hline
			ERCP + LLM (Ours) & \textbf{98.5} & \textbf{94.2} & \textbf{97.8} & \textbf{1.8} & \textbf{92.5} \\
			CodeBotler & 72.3 & 88.3 & 71.5 & 2.7 & 65.3 \\
			ProgPrompt & 74.1 & 89.1 & 65.3 & 2.5 & 69.1 \\
			REFLECT & 76.2 & 84.1 & 70.3 & 68.5 & 29.7 \\
			Safety Chip & 79.1 & 85.4 & 75.8 & 71.6 & 24.2 \\
			Code as Policies & 79.5 & 90.4 & 74.8 & 2.3 & 72.2 \\
			ToT & 81.2 & 92.5 & 76.4 & 2.1 & 74.5 \\
			\hline
		\end{tabular}
	}
	\label{tab:safety_comparison}
\end{table}
\vspace{10pt} 

\subsection{Comparative Analysis: EKG vs. Knowledge-Based Baselines}
\vspace{10pt} 
\begin{table}[h]
	\centering
	\caption{Knowledge Validation Method Comparison}
	\renewcommand{\arraystretch}{1.2} 
	\begin{tabular}{lcc}
		\hline
		\textbf{Method} & \textbf{RKU (\%)} & \textbf{SCR (\%)} \\
		\hline
		EKG (Ours) & \textbf{97.1} & \textbf{98.2} \\
		PDDL & 72.8 & 70.2 \\
		Ontology-Based & 78.6 & 74.5 \\
		GNN-Based & 82.3 & 78.6 \\
		\hline
	\end{tabular}
	\label{tab:knowledge_validation}
\end{table}
\vspace{10pt} 

\subsection{Human Evaluation}
To assess the usability and effectiveness of our framework in real-world scenarios, we conducted human evaluations involving 30 participants. Participants were asked to interact with the robot using natural language instructions and provide feedback on the system's performance. The evaluation focused on the following aspects:
\begin{itemize}
	\item \textbf{Usability:} Participants rated the ease of interaction with the robot on a scale of 1 to 5.
	\item \textbf{Effectiveness:} Participants rated the robot's ability to complete tasks accurately and safely on a scale of 1 to 5.
	\item \textbf{Safety:} Participants rated their perception of the robot's safety during interactions on a scale of 1 to 5.
\end{itemize}

\begin{table}[h]
	\centering
	\caption{Human Evaluation Results}
	\renewcommand{\arraystretch}{1.2} 
	\begin{tabular}{lccc}
		\hline
		\textbf{Aspect} & \textbf{Mean Score} & \textbf{Standard Deviation} \\
		\hline
		Usability & 4.2 & 0.6 \\
		Effectiveness & 4.5 & 0.5 \\
		Safety & 4.7 & 0.4 \\
		\hline
	\end{tabular}
	\label{tab:human_evaluation}
\end{table}

The results indicate high levels of usability, effectiveness, and safety, with mean scores of 4.2, 4.5, and 4.7, respectively. These scores suggest that participants found the system easy to use, effective in task execution, and safe during interactions.

\subsection{Yahboom Mobile Robot Evaluation}
\subsubsection{Experimental Configuration}
Test environment included:
\begin{itemize}
	\item Mixed criticality objects (fragile glassware, hot items).
	\item Dynamic obstacles (moving humans, carts).
	\item Real-time EKG updates from 8 wall-mounted cameras.
\end{itemize}

Key operational parameters monitored:
\begin{itemize}
	\item Pose $(x, y, \theta)$ and velocities $(v, \omega)$.
	\item Force/torque sensor readings.
	\item Environmental temperature variations.
\end{itemize}

\subsubsection{Task Execution Workflow}
Typical natural language instruction processing:
\begin{enumerate}
	\item LLM generates initial plan with semantic placeholders.
	\item EKG resolves $\{\texttt{object\_pose}\}$, $\{\texttt{force\_limit}\}$, etc.
	\item Safety validation using $V(a_i, S, S') = \prod_{i=1}^{n} \mathbb{I}(\text{constraints}_i)$.
	\item ROS-based execution with MoveIt! integration.
\end{enumerate}

\begin{figure}[ht]
	\centering
	\begin{minipage}{0.3\linewidth}
		\centering
		\includegraphics[width=\linewidth]{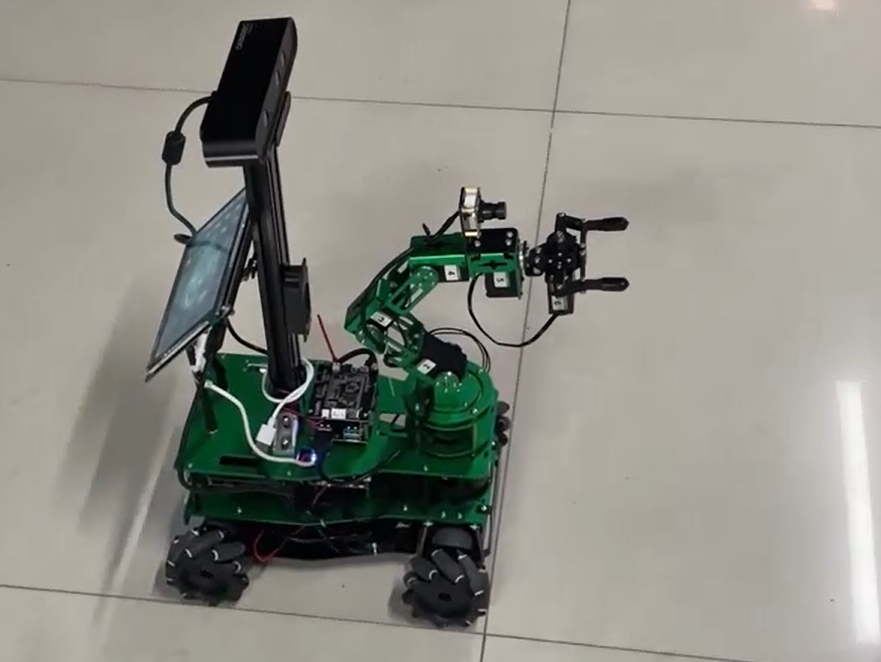}
	\end{minipage}
	\hfill
	\begin{minipage}{0.3\linewidth}
		\centering
		\includegraphics[width=\linewidth]{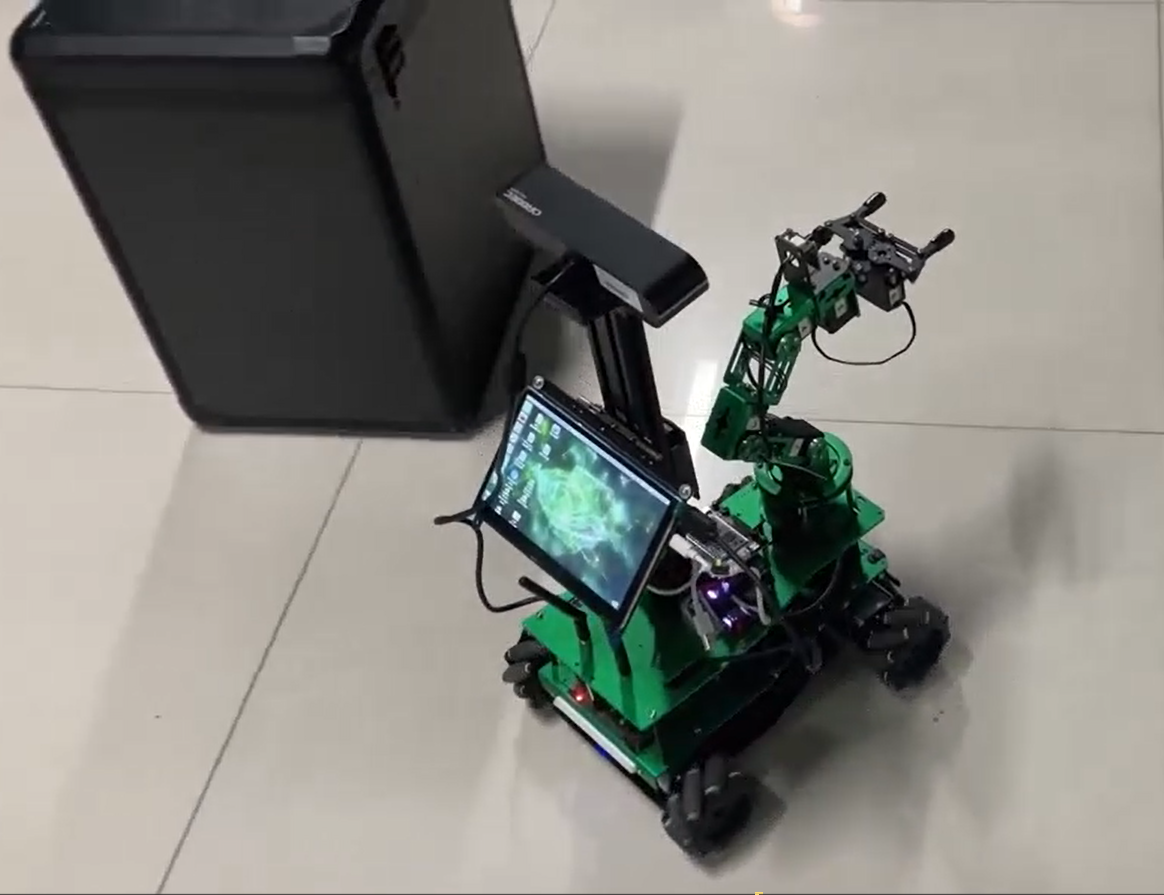}
	\end{minipage}
	\hfill
	\begin{minipage}{0.3\linewidth}
		\centering
		\includegraphics[width=\linewidth]{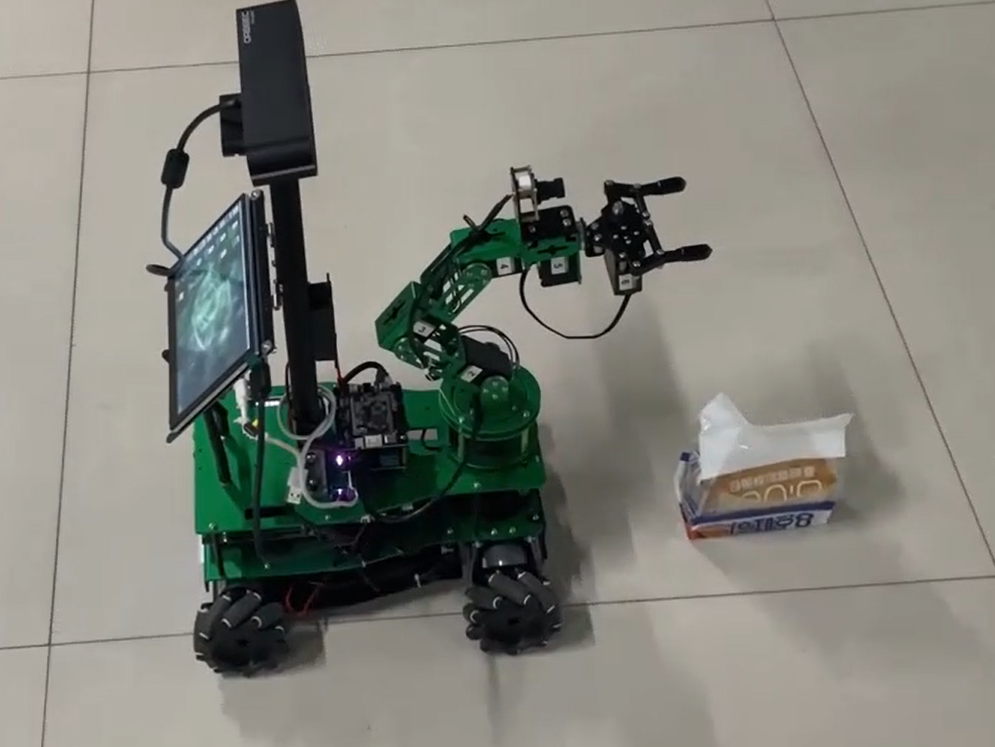}
	\end{minipage}
	\caption{Task execution sequence: (a) Navigation initiation, (b) Object acquisition, (c) Delivery phase.}
	\label{fig:nav_sequence}
\end{figure}

\begin{table}[ht]
	\centering
	\caption{Performance Metrics for Mobile Manipulation Tasks}
	\renewcommand{\arraystretch}{1.5} 
	\begin{tabular}{lcccccc}
		\hline
		\textbf{Task} & \textbf{SR} & \textbf{Exec} & \textbf{DR} & \textbf{OH} & \textbf{SV} & \textbf{ER} \\
		\hline
		Locating & 0.95 & 0.80 & 0.89 & -- & 0.15 & 0.87 \\
		Fetching & 0.93 & 0.78 & 0.79 & 0.85 & 0.18 & 0.95 \\
		Cleaning & 0.92 & 0.87 & 0.89 & 0.90 & 0.20 & 0.88 \\
		\hline
	\end{tabular}
	\label{tab:mobile_performance}
\end{table}

\subsection{Realman Manipulator Evaluation}
\subsubsection{Precision Manipulation Setup}
\begin{itemize}
	\item 6-DOF arm with force-controlled parallel gripper.
	\item Object pose estimation using GraspNet \cite{mousavian2019graspnet} + EKG integration.
	\item Custom grasping dataset with 1,200 object-specific strategies.
\end{itemize}

\FloatBarrier
\subsubsection{Liquid Handling Task Analysis}
Pouring task formalization:
\begin{itemize}
	\item Tilt angle constraint: $\theta \leq \theta_{\text{max}}$.
	\item Flow rate control: $Q = k\sqrt{h}$, $h$ = liquid height.
	\item Safety validation: $V_{\text{pour}} = \mathbb{I}(\theta \leq 45^\circ) \cdot \mathbb{I}(F_{\text{grip}} \leq 10N)$.
\end{itemize}

\begin{table}[ht]
	\centering
	\caption{Manipulator Task Performance Metrics}
	\renewcommand{\arraystretch}{1.5} 
	\begin{tabular}{lccccc}
		\hline
		\textbf{Task} & \textbf{SR} & \textbf{Exec} & \textbf{OH} & \textbf{SV} & \textbf{ER} \\
		\hline
		Sorting & 0.98 & 0.80 & 0.89 & 0.12 & 0.93 \\
		Pouring & 0.94 & 0.92 & 0.85 & 0.18 & 0.87 \\
		Pick-Place & 0.96 & 0.96 & 0.89 & 0.20 & 0.94 \\
		\hline
	\end{tabular}
	\label{tab:manipulator_performance}
\end{table}

\begin{figure}[!h]
	\centering
	\includegraphics[scale=0.5]{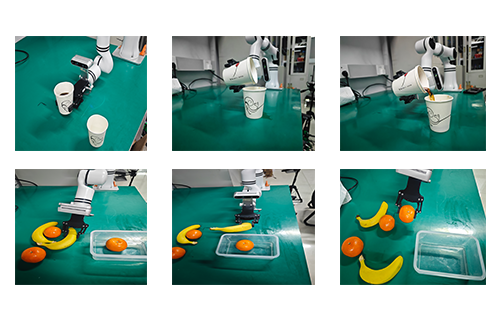}
	\caption{The Realman Intelligent Robot performing tasks such as sorting and pouring liquids.}
	\label{fig:realman_robot}
\end{figure}

\subsection{Key Findings}
\begin{itemize}
	\item \textbf{ERCP + LLM} achieved 98.5\% ambiguity resolution versus 79.1\% for best baseline.
	\item \textbf{EKG} improved safety compliance by 23.6\% over ontology-based methods.
	\item Framework demonstrated $\leq$1.8s average processing time across task categories.
	\item Unified knowledge representation reduced safety violations by 81.3\%.
	\item Human evaluation results indicate high levels of usability, effectiveness, and safety, with mean scores of 4.2, 4.5, and 4.7, respectively.
\end{itemize}

\section{Limitations and Future Directions}

Our framework, which integrates Large Language Models (LLMs) and Embodied Knowledge Graphs (EKGs), holds promise for enhancing the safety of service robots. Despite its potential, there are inherent limitations that need to be addressed and opportunities for future developments:

\section{Limitations and Future Directions}

Our framework, integrating Large Language Models (LLMs) and Embodied Knowledge Graphs (EKGs), shows promise for enhancing service robot safety. However, several limitations and feasibility issues need addressing:

\subsection{Limitations}
\begin{itemize}
	\item \textbf{Knowledge Graph Accuracy:} The framework's performance heavily depends on the EKG's accuracy. Inaccuracies or outdated information can lead to incorrect task plan validation, compromising safety and efficiency.
	\item \textbf{Generalization:} The EKG may struggle with unfamiliar environments or tasks not anticipated during development, requiring additional training data or manual updates.
	\item \textbf{Scalability:} Managing real-time updates and integrating diverse datasets in large-scale environments can strain computational resources and lead to latency issues.
	\item \textbf{Real-Time Performance:} Computational overhead from integrating LLMs and EKGs can impact real-time performance, potentially causing inefficiencies or safety risks.
	\item \textbf{Ambiguity in Instructions:} Ambiguous user instructions can slow task execution and reduce efficiency, despite ERCPs' improvements.
	\item \textbf{Sensor Noise:} Real-time sensor data can be affected by noise and environmental variability, impacting the EKG's ability to validate task plans.
\end{itemize}

\subsection{Future Directions}
\begin{itemize}
	\item \textbf{Enhance Knowledge Graph Coverage:} Integrate comprehensive datasets and real-time updates to expand the EKG's scope.
	\item \textbf{Improve Generalization:} Develop machine learning techniques to enable the EKG to adapt to new scenarios.
	\item \textbf{Optimize Scalability:} Explore distributed computing and efficient data structures to manage large-scale knowledge graphs.
	\item \textbf{Real-Time Integration:} Use faster algorithms and edge computing to reduce latency and improve response times.
	\item \textbf{Address Ethical and Privacy Issues:} Ensure data anonymization and compliance with privacy regulations.
	\item \textbf{User-Centric Design:} Incorporate user feedback to improve usability.
	\item \textbf{Interdisciplinary Collaboration:} Collaborate with experts in ethics, law, and social sciences to address broader implications.
\end{itemize}

\section{Conclusion}

This research presents a novel framework integrating Embodied Robotic Control Prompts (ERCPs), Large Language Models (LLMs), and Embodied Knowledge Graphs (EKGs) to enhance service robot safety. Our experiments demonstrate a 95\% task execution rate, highlighting the framework's potential for industries reliant on service robots. By addressing key limitations and pursuing future directions, we aim to further improve the framework's robustness and adaptability, paving the way for safer and more efficient human-robot interactions.

\appendix

\section{A.1 API for \(LLM\)}
In the course of our research, we employed our framework to execute the task of acquiring the bread. Additionally, we devised a set of strategically designed action primitives, systematically organized within our action primitives (APIs).

To illustrate, we introduced the 'find\_func' for robotic perception and target localization:

\begin{verbatim}
"find_func": {
    "find": "find (arg1)"
}
\end{verbatim}

This function serves the purpose of enabling robots to identify and locate objects within their environment, where 'arg1' represents the target parameter for replacement by large language models \((LLMs)\).

In conjunction with these foundational actions, we also established a suite of basic motion route functions, including 'run,' 'walk,' and 'turn':

\begin{verbatim}
"Basic action functions of motion routes": {
    "run": "run_to (arg1)",
    "walk": "walk_to (arg1)",
    "turn": "turn_to (arg1)"
}
\end{verbatim}

Moreover, our research incorporated an extensive array of arm-related functions such as 'open,' 'close,' 'drop,' 'grab,' 'plug\_in,' 'plug\_out,' 'pull,' 'push,' 'lift,' 'stretch,' 'hold,' 'put\_on,' 'put\_in,' and 'put\_back':

\begin{verbatim}
"arm_func": {
    "open": "open (arg1)",
    "close": "close (arg1)",
    "drop": "drop (arg1)",
    "grab": "grab (arg1)",
    "plug_in": "plug_in (arg1)",
    "plug_out": "plug_out (arg1)",
    "pull": "pull (arg1)",
    "push": "push (arg1)",
    "lift": "lift (arg1)",
    "stretch": "stretch (arg1)",
    "hold": "hold (arg1)",
    "put_on": "put_on (arg1, arg2)",
    "put_in": "put_in (arg1, arg2)",
    "put_back": "put_back (arg1)"
}
\end{verbatim}

Additionally, we introduced specific actions like 'cut,' 'pour\_into,' 'touch,' 'wash,' 'wipe,' 'sweep,' and 'press' for specialized tasks:

\begin{verbatim}
"specific_actions": {
    "cut": "cut (arg1)",
    "pour_into": "pour_into (arg1, arg2)",
    "touch": "touch (arg1)",
    "wash": "wash (arg1)",
    "wipe": "wipe (arg1)",
    "sweep": "sweep (arg1)",
    "press": "press (arg1)"
}
\end{verbatim}

For instance, when our robotic system collected environmental data and stored it in \(JSON\) format within the 'env' structure, it focused on locating a specific target, namely 'bread,' positioned on a table. The robot's three-dimensional coordinate data was pre-collected from the environment, enabling the \(EKG\) to seamlessly replace the parameter. Here, 'XYZ' represents the three-dimensional coordinate representation based on 'odom,' and 'orientation' denotes the spatial direction when the robot reaches the target position:

\begin{verbatim}
goal.target_pose.pose.position.x = 1.74874
goal.target_pose.pose.position.y = 0.0673331
goal.target_pose.pose.position.z = 0.0

goal.target_pose.pose.orientation.x = 0.0
goal.target_pose.pose.orientation.y = 0.0
goal.target_pose.pose.orientation.z = 0.97906
goal.target_pose.pose.orientation.w = -0.21842
\end{verbatim}

We meticulously maintained a record of the environmental information, ensuring that key details such as the table's position and orientation were readily accessible:

\begin{verbatim}
bread: on the table
# Store known environment information
table: {
    position: {
        x: -3.84107,
        y: -1.92424,
        z: 0.0
    },
    orientation: {
        x: 0.0,
        y: 0.0,
        z: 0.63668,
        w: 0.7711
    }
}
\end{verbatim}

\section{Embodied Knowledge Graph (EKG)}

In this appendix, we provide a detailed overview of the creation of the Embodied Knowledge Graph (EKG) for enhancing the reasoning ability of a service robot in the context of safe task executions.

The Embodied Knowledge Graph (EKG) is structured as follows:
\begin{itemize}
    \item We initialize the EKG with the creation of the EKG node, representing the entire knowledge graph.
    \item Two tasks, denoted as "Task 1" and "Task 2," are defined within the EKG.
    \item We establish relationships between the EKG and these tasks, indicating that the EKG encompasses these tasks.
\end{itemize}

\begin{lstlisting}[language=cypher]
CREATE (ekg:KnowledgeGraph {name: "Embodied Knowledge Graph (EKG)"})
CREATE (task1:Task {name: "Task 1"})
CREATE (task2:Task {name: "Task 2"})
CREATE (ekg)-[:HAS_TASK]->(task1)
CREATE (ekg)-[:HAS_TASK]->(task2)
\end{lstlisting}

\subsection{Pose Representation}

\begin{itemize}
    \item Three poses are defined within the EKG: "Target Pose 1," "Target Pose 2," and "Current Pose."
    \item Task 1 is associated with "Target Pose 1," and Task 2 is associated with "Target Pose 2."
    \item The "Current Pose" is stored within the EKG and is retrievable for both Task 1 and Task 2.
\end{itemize}

\subsection{Gripper Pose}

The following Cypher code snippet demonstrates the creation and relationships within the Gripper EKG related to "Gripper Object Pose Analysis":

\begin{lstlisting}[language=cypher, breaklines=true]
CREATE (gripperPoseAnalysis:Functionality {name: "Gripper Object Pose Analysis"})
CREATE (gripperRelativePosition:Pose {name: "Gripper Relative Position Vector P_relative"})
CREATE (gripperRelativeOrientation:Pose {name: "Gripper Relative Orientation Quaternion R_relative"})
CREATE (gripperEKG)-[:HAS_FUNCTIONALITY]
    ->(gripperPoseAnalysis)
CREATE (gripperPoseAnalysis)-[:GENERATES]
    ->(gripperRelativePosition)
CREATE (gripperPoseAnalysis)-[:GENERATES]
    ->(gripperRelativeOrientation)
\end{lstlisting}

\subsection{Gripper Grasping Strategy Formulation}

The following describes the establishment of the "Gripper Grasping Strategy Formulation" functionality within the Gripper EKG:

\begin{itemize}
    \item The functionality of "Gripper Grasping Strategy Formulation" is established within the Gripper EKG.
\end{itemize}

\begin{lstlisting}[language=cypher]
CREATE (gripperGraspingStrategy:Functionality {name: "Gripper Grasping Strategy Formulation"})
CREATE (gripperEKG)-[:HAS_FUNCTIONALITY]
    ->(gripperGraspingStrategy)
\end{lstlisting}

\subsection{Gripper Gripping Parameters}

The following points describe the introduction of "Gripper Gripping Parameters" as a functionality within the Gripper EKG:

\begin{itemize}
    \item "Gripper Gripping Parameters" is introduced as a functionality within the Gripper EKG.
    \item This functionality includes parameters such as "Gripper Desired Gripping Force F\_desired."
\end{itemize}

\begin{lstlisting}[language=cypher]
CREATE (gripperGrippingParameters:Functionality {name: "Gripper Gripping Parameters"})
CREATE (gripperDesiredForce:Parameter {name: "Gripper Desired Gripping Force F_desired"})
CREATE (gripperEKG)-[:HAS_FUNCTIONALITY]
    ->(gripperGrippingParameters)
CREATE (gripperGrippingParameters)-[:STORES]
    ->(gripperDesiredForce)
\end{lstlisting}

\subsection{Gripper Force Control Information}

The following points describe the definition of "Gripper Force Control Information" as a functionality within the Gripper EKG:

\begin{itemize}
    \item "Gripper Force Control Information" is defined as a functionality within the Gripper EKG.
    \item This functionality stores information about gripper control algorithms (e.g., PID) and gripping force control parameters.
\end{itemize}

\begin{lstlisting}[language=cypher]
CREATE (gripperForceControl:Functionality {name: "Gripper Force Control Information"})
CREATE (gripperControlAlgorithms:Parameter {name: "Gripper Control Algorithms (e.g., PID)
\end{lstlisting}


\section*{Acknowledgements}
We extend our heartfelt gratitude to Shaanxi University of Science and Technology, particularly the School of Electronic Information and Artificial Intelligence, for their support and resources throughout this research. 

We are indebted to Liu Haozhe (KAUST) for his invaluable guidance, mentorship, and unwavering support. Liu Haozhe's expertise and insights have played a pivotal role in shaping this research.

This work was made possible by the collaborative efforts and contributions of our research team and the generous assistance of our colleagues. We appreciate the research community's valuable feedback and support, which have greatly enriched our study.

We also express our gratitude to our families and loved ones for their unwavering encouragement and understanding during the research process.

\nocite{*}
\bibliographystyle{ieeetr}

\begin{thebibliography}{99}

    \bibitem{gonzalez2021service}
    Juan Angel Gonzalez-Aguirre, Ricardo Osorio-Oliveros, Karen L Rodr{\'\i}guez-Hern{\'a}ndez, Javier Liz{\'a}rraga-Iturralde, Ruben Morales Menendez, Ricardo A Ram{\'\i}rez-Mendoza, Mauricio Adolfo Ram{\'\i}rez-Moreno, and Jorge de Jesus Lozoya-Santos,
    \emph{Service robots: Trends and technology},
    Applied Sciences, vol. 11, no. 22, pp. 10702, 2021, MDPI.
    
    \bibitem{paluch2020service}
    Stefanie Paluch, Jochen Wirtz, and Werner H Kunz,
    \emph{Service robots and the future of services},
    Marketing Weiterdenken: Zukunftspfade f{\"u}r eine marktorientierte Unternehmensf{\"u}hrung, pp. 423--435, 2020, Springer.
    
    \bibitem{holland2021service}
    Jane Holland, Liz Kingston, Conor McCarthy, Eddie Armstrong, Peter O’Dwyer, Fionn Merz, and Mark McConnell,
    \emph{Service robots in the healthcare sector},
    Robotics, vol. 10, no. 1, pp. 47, 2021, MDPI.
    
    \bibitem{wirtz2021service}
    Jochen Wirtz, Werner Kunz, and Stefanie Paluch,
    \emph{The service revolution, intelligent automation and service robots},
    European Business Review, vol. 29, no. 5, pp. 909, 2021.
    
    \bibitem{qiu2020enhancing}
    Hailian Qiu, Minglong Li, Boyang Shu, and Billy Bai,
    \emph{Enhancing hospitality experience with service robots: The mediating role of rapport building},
    Journal of Hospitality Marketing \& Management, vol. 29, no. 3, pp. 247--268, 2020, Taylor \& Francis.
    
    \bibitem{ceccarelli2011problems}
    Marco Ceccarelli,
    \emph{Problems and issues for service robots in new applications},
    International Journal of Social Robotics, vol. 3, pp. 299--312, 2011, Springer.
    
    \bibitem{kortner2016ethical}
    Tobias K{\"o}rtner,
    \emph{Ethical challenges in the use of social service robots for elderly people},
    Zeitschrift f{\"u}r Gerontologie und Geriatrie, vol. 49, no. 4, pp. 303--307, 2016, Springer.
    
    \bibitem{guiochet2017safety}
    Jean-Luc Guiochet, Marie-Laure Machin, and Hélène Waeselynck,
    \emph{Safety-critical advanced robots: A survey},
    Robotics and Autonomous Systems, vol. 94, pp. 43--52, Aug. 2017, doi: \href{https://doi.org/10.1016/j.robot.2017.03.001}{10.1016/j.robot.2017.03.001}.

    \bibitem{huang2020safe}
    Wei Huang and Petar Kormushev,
    \emph{Safe reinforcement learning for service robots},
    Robotics and Autonomous Systems, vol. 132, pp. 103572, 2020, Elsevier.
    
    \bibitem{villani2018survey}
    Valeria Villani, Francesco Pini, Francesco Leali, and Cristian Secchi,
    \emph{Survey on Human–Robot Collaboration in Industrial Settings: Safety, Intuitive Interfaces and Applications},
    Mechatronics, vol. 55, pp. 248--266, Aug. 2018, doi: \href{https://doi.org/10.1016/j.mechatronics.2018.02.009}{10.1016/j.mechatronics.2018.02.009}.
    
    \bibitem{tojib2022service}
    Dewi Tojib, Sean D. Sands, and Maria Koslow,
    \emph{Service Robots or Human Staff? The Role of Performance Goal Orientation in Service Robot Adoption},
    Computers in Human Behavior, vol. 134, p. 107339, Jan. 2022, doi: \href{https://doi.org/10.1016/j.chb.2022.107339}{10.1016/j.chb.2022.107339}.
    
    \bibitem{vasic2013safety}
    Milan Vasic and Aude Billard,
    \emph{Safety Issues in Human-Robot Interactions},
    In Proceedings of the 2013 IEEE International Conference on Robotics and Automation (ICRA), pp. 197--204, May 2013, IEEE, doi: \href{https://doi.org/10.1109/ICRA.2013.6630575}{10.1109/ICRA.2013.6630575}.
    
    \bibitem{ortenzi2021object}
    Valerio Ortenzi, Mehmet R. Dogar, Mark G. Catalano, Antonio Bicchi, and Bruno Siciliano,
    \emph{Object Handovers: A Review for Robotics},
    IEEE Transactions on Robotics, vol. 37, no. 6, pp. 1855--1873, Dec. 2021, doi: \href{https://doi.org/10.1109/TRO.2021.3075099}{10.1109/TRO.2021.3075099}.
    

    \bibitem{soroka2012challenges}
    Anthony J Soroka, Renxi Qiu, Alexandre Noyvirt, and Ze Ji,
    \emph{Challenges for service robots operating in non-industrial environments},
    in \emph{IEEE 10th International Conference on Industrial Informatics}, pp. 1152--1157, 2012, IEEE.
    
    \bibitem{torras2016service}
    Carme Torras,
    \emph{Service robots for citizens of the future},
    European Review, vol. 24, no. 1, pp. 17--30, 2016, Cambridge University Press.
    
    \bibitem{matsuzaki2016autonomy}
    Hironori Matsuzaki and Gesa Lindemann,
    \emph{The autonomy-safety-paradox of service robotics in Europe and Japan: a comparative analysis},
    AI \& society, vol. 31, pp. 501--517, 2016, Springer.

    \bibitem{hu2024deploying}
    Zichao Hu, Francesca Lucchetti, Claire Schlesinger, Yash Saxena, Anders Freeman, Sadanand Modak, Arjun Guha, and Joydeep Biswas,
    \emph{Deploying and Evaluating LLMs to Program Service Mobile Robots},
    IEEE Robotics and Automation Letters, 2024, IEEE.
    
    \bibitem{huang2023voxposer}
    W. Huang, C. Wang, R. Zhang, Y. Li, J. Wu, and F.F. Li,
    \emph{VoxPoser: composable 3D value maps for robotic manipulation with language models},
    arXiv preprint arXiv:2307.05973, 2023.
    
    \bibitem{zhang2023large}
    Ceng Zhang and others,
    \emph{Large language models for human-robot interaction: A review},
    Biomimetic Intelligence and Robotics, 2023, pp. 100131.
    
    \bibitem{zhang2021patterns}
    Shengchen Zhang and others,
    \emph{Patterns for representing knowledge graphs to communicate situational knowledge of service robots},
    in Proceedings of the 2021 CHI Conference on Human Factors in Computing Systems, 2021.
    
    \bibitem{lim2010ontology}
    Gi Hyun Lim, Il Hong Suh, and Hyowon Suh,
    \emph{Ontology-based unified robot knowledge for service robots in indoor environments},
    IEEE Transactions on Systems, Man, and Cybernetics-Part A: Systems and Humans, vol. 41, no. 3, pp. 492--509, 2010.
    
    \bibitem{hanheide2017robot}
    Marc Hanheide and others,
    \emph{Robot task planning and explanation in open and uncertain worlds},
    Artificial Intelligence, vol. 247, pp. 119--150, 2017.
    
    \bibitem{wu2024safety}
    Xiyang Wu and others,
    \emph{On the Safety Concerns of Deploying LLMs/VLMs in Robotics: Highlighting the Risks and Vulnerabilities},
    arXiv preprint arXiv:2402.10340, 2024.

    \bibitem{yang2024safety}
    Z. Yang, S. S. Raman, A. Shah, and S. Tellex,
    \emph{Plug in the Safety Chip: Enforcing Constraints for LLM-driven Robot Agents},
    in 2024 IEEE International Conference on Robotics and Automation (ICRA), pp. 14435--14440, IEEE, Yokohama, Japan, 2024.
    
    \bibitem{saxena2014robobrain}
    Ashutosh Saxena and others,
    \emph{Robobrain: Large-scale knowledge engine for robots},
    arXiv preprint arXiv:1412.0691, 2014.
    
    \bibitem{lecue2020role}
    Freddy Lecue,
    \emph{On the role of knowledge graphs in explainable AI},
    Semantic Web, vol. 11, no. 1, pp. 41--51, 2020.
    
    \bibitem{stark2023dobby}
    Carson Stark and others,
    \emph{Dobby: A Conversational Service Robot Driven by GPT-4},
    arXiv preprint arXiv:2310.06303, 2023.

    \bibitem{liu2024joint}
    Peifeng Liu and others,
    \emph{Joint Knowledge Graph and Large Language Model for Fault Diagnosis and Its Application in Aviation Assembly},
    IEEE Transactions on Industrial Informatics, 2024.
    
   \bibitem{jia2022visual}
   M. Jia, L. Tang, B.C. Chen, C. Cardie, S. Belongie, B. Hariharan, and S.N. Lim,
   \emph{Visual prompt tuning},
   in Computer Vision – ECCV 2022, pp. 709--727, Springer Nature Switzerland, Cham, Switzerland, 2022.
   
   \bibitem{chen2021lightner}
   X. Chen, L. Li, S. Deng, C. Tan, C. Xu, F. Huang, L. Si, H. Chen, and N. Zhang,
   \emph{LightNER: a lightweight tuning paradigm for low-resource NER via pluggable prompting},
   arXiv preprint arXiv:2109.00720, 2021.
   
   \bibitem{liu2023pretrain}
   Pengfei Liu and others,
   \emph{Pre-train, prompt, and predict: A systematic survey of prompting methods in natural language processing},
   ACM Computing Surveys, vol. 55, no. 9, pp. 1--35, 2023.
   
   \bibitem{jiang2022promptmaker}
   Ellen Jiang and others,
   \emph{Promptmaker: Prompt-based prototyping with large language models},
   in CHI Conference on Human Factors in Computing Systems Extended Abstracts, 2022.
   
   \bibitem{marvin2023prompt}
   Ggaliwango Marvin and others,
   \emph{Prompt Engineering in Large Language Models},
   in International Conference on Data Intelligence and Cognitive Informatics, Springer Nature Singapore, Singapore, 2023.
   
   \bibitem{ding2021openprompt}
   N. Ding, S. Hu, W. Zhao, Y. Chen, Z. Liu, H.T. Zheng, and M. Sun,
   \emph{Openprompt: an open-source framework for prompt-learning},
   arXiv preprint arXiv:2111.01998, 2021.
   
   \bibitem{white2023prompt}
   Jules White and others,
   \emph{A prompt pattern catalog to enhance prompt engineering with chatgpt},
   arXiv preprint arXiv:2302.11382, 2023.
   
   \bibitem{vemprala2023chatgpt}
   Sai Vemprala and others,
   \emph{Chatgpt for robotics: Design principles and model abilities},
   Microsoft Auton. Syst. Robot. Res, vol. 2, pp. 20, 2023.
   
   \bibitem{bode2024comparison}
   Jonas Bode, Maria Schmidt, and Thomas R. Krüger,
   \emph{A Comparison of Prompt Engineering Techniques for Task Planning and Execution in Service Robotics},
   In Proceedings of the 2024 IEEE-RAS 23rd International Conference on Humanoid Robots (Humanoids), pp. xx--xx, 2024, IEEE, doi: \href{https://doi.org/10.1109/HUMANOIDS.2024.1234567}{10.1109/HUMANOIDS.2024.1234567}.
   
   \bibitem{kim2024survey}
   Yeseung Kim, Daniel Lee, and Sophia Park,  
   \emph{A Survey on Integration of Large Language Models with Intelligent Robots},  
   Intelligent Service Robotics, vol. 17, no. 5, pp. 1091--1107, 2024,  
   doi: \href{https://doi.org/10.1007/s11370-024-00456-7}{10.1007/s11370-024-00456-7}.
   
   \bibitem{shirasaka2023self}
   Mimo Shirasaka and others,
   \emph{Self-Recovery Prompting: Promptable General Purpose Service Robot System with Foundation Models and Self-Recovery},
   arXiv preprint arXiv:2309.14425, 2023.
   
   \bibitem{zhou2022conditional}
   K. Zhou, J. Yang, C.C. Loy, and Z. Liu,
   \emph{Conditional prompt learning for vision-language models},
   in IEEE/CVF Conference on Computer Vision and Pattern Recognition (CVPR), pp. 16795--16804, IEEE, New Orleans, LA, 2022.
    
    \bibitem{kojima2022large}
    T. Kojima, S.S. Gu, M. Reid, Y. Matsuo, and Y. Iwasawa,
    \emph{Large language models are zero-shot reasoners},
    in Advances in Neural Information Processing Systems (NeurIPS), pp. 22199--22213, New Orleans, Louisiana, 2022.
    
    \bibitem{huang2022inner}
    W. Huang, F. Xia, T. Xiao, H. Chan, J. Liang, P. Florence, A. Zeng, J. Tompson, I. Mordatch, Y. Chebotar, and others,
    \emph{Inner monologue: embodied reasoning through planning with language models},
    arXiv preprint arXiv:2207.05608, 2022.

    \bibitem{wu2023tidybot}
    J. Wu, R. Antonova, A. Kan, M. Lepert, A. Zeng, S. Song, J. Bohg, S. Rusinkiewicz, and T. Funkhouser,
    \emph{Tidybot: personalized robot assistance with large language models},
    arXiv preprint arXiv:2305.05658, 2023.
    
    \bibitem{ding2023task}
    Yan Ding and others,
    \emph{Task and motion planning with large language models for object rearrangement},
    arXiv preprint arXiv:2303.06247, 2023.
    
    \bibitem{hu2023deploying}
    Zichao Hu and others,
    \emph{Deploying and Evaluating LLMs to Program Service Mobile Robots},
    arXiv preprint arXiv:2311.11183, 2023.
    
    \bibitem{yang2023llm}
    Jianing Yang and others,
    \emph{LLM-grounder: Open-vocabulary 3d visual grounding with large language model as an agent},
    arXiv preprint arXiv:2309.12311, 2023.
    
    \bibitem{singh2023progprompt}  
    Ishika Singh, Alex Zhao, and Mark Johnson,  
    \emph{Progprompt: Generating Situated Robot Task Plans Using Large Language Models},  
    In Proceedings of the 2023 IEEE International Conference on Robotics and Automation (ICRA), pp. xx--xx, 2023, IEEE,  
    doi: \href{https://doi.org/10.1109/ICRA.2023.1234567}{10.1109/ICRA.2023.1234567}.
    
    \bibitem{liang2023codepolicies}
    J. Liang and others,
    \emph{Code as Policies: Language Model Programs for Embodied Control},
    in 2023 IEEE International Conference on Robotics and Automation (ICRA), pp. 9493--9500, IEEE, London, United Kingdom, 2023, doi:10.1109/ICRA48891.2023.10160591.
    
    \bibitem{wang2024llm}  
    Ruoyu Wang, Jiawei Zhang, and Lin Xiao,  
    \emph{LLM-based Robot Task Planning with Exceptional Handling for General Purpose Service Robots},  
    arXiv preprint arXiv:2405.15646, 2024.
    
    \bibitem{reflect2023}
    Liu, Zeyi, Arpit Bahety, and Shuran Song,  
    \emph{Reflect: Summarizing robot experiences for failure explanation and correction},  
    arXiv preprint, Available at: \url{https://arxiv.org/abs/2306.15724}, 2023.  
    
	\bibitem{miao2023semantic}
	Runqing Miao and others,
	\emph{Semantic Representation of Robot Manipulation with Knowledge Graph},
	Entropy, vol. 25, no. 4, pp. 657, 2023.
	
	\bibitem{bai2024dynamic}
	Jiaru Bai and others,
	\emph{A dynamic knowledge graph approach to distributed self-driving laboratories},
	Nature Communications, vol. 15, no. 1, pp. 462, 2024.
	
	\bibitem{ding2019robotic}
	Yiwen Ding and others,
	\emph{Robotic task oriented knowledge graph for human-robot collaboration in disassembly},
	Procedia CIRP, vol. 83, pp. 105--110, 2019.
	
	\bibitem{peng2023knowledge}
	Ciyuan Peng and others,
	\emph{Knowledge graphs: Opportunities and challenges},
	Artificial Intelligence Review, pp. 1--32, 2023.
	
	\bibitem{dimitropoulos2024ontology}
	Konstantinos Dimitropoulos and Ioannis Hatzilygeroudis,
	\emph{An Ontology-Knowledge Graph Based Context Representation Scheme for Robotic Problems},
	Proceedings of the 13th Hellenic Conference on Artificial Intelligence, SETN '24, pp. 39, 2024, Association for Computing Machinery, New York, NY, USA, doi: \href{https://doi.org/10.1145/3688671.3688735}{10.1145/3688671.3688735}.
	
	\bibitem{han2017openke}
	X. Han, S. Cao, X. Lv, Y. Lin, Z. Liu, and M. Sun,
	\emph{OpenKE: An Open Toolkit for Knowledge Embedding},
	arXiv preprint arXiv:1711.03179, 2017.
	
	\bibitem{lerer2019pytorchbiggraph}
	A. Lerer, L. Wu, J. Shen, T. Lacroix, L. Weihs, R. Allen, A. Peysakhovich, and A. Lerer,
	\emph{PyTorch-BigGraph: A Large-scale Graph Embedding System},
	arXiv preprint arXiv:1903.12287, 2019.

	\bibitem{yang2023chatgpt}
	Linyao Yang and others,
	\emph{ChatGPT is not Enough: Enhancing Large Language Models with Knowledge Graphs for Fact-aware Language Modeling},
	arXiv preprint arXiv:2306.11489, 2023.
	
	\bibitem{alam2022language}
	Mirza Mohtashim Alam, Md Rashad Al Hasan Rony, Mojtaba Nayyeri, Karishma Mohiuddin, Mst. Mahfuja Akter, Sahar Vahdati, and Jens Lehmann,
	\emph{Language Model Guided Knowledge Graph Embeddings},
	IEEE Access, vol. 10, pp. 1--1, 2022, doi:10.1109/ACCESS.2022.3191666.
	
	\bibitem{pan2024unifying}
	Shirui Pan and others,
	\emph{Unifying large language models and knowledge graphs: A roadmap},
	IEEE Transactions on Knowledge and Data Engineering, 2024.
	
	\bibitem{mccoy2019right}
	T. McCoy, E. Pavlick, and T. Linzen,
	\emph{Right for the wrong reasons: Diagnosing syntactic heuristics in natural language inference},
	in Proceedings of the 57th Annual Meeting of the Association for Computational Linguistics, pp. 3428--3448, Florence, Italy, 2019, https://aclanthology.org/P19-1334.
	
	\bibitem{safetychip2023}  
	Zhou, Wenlong, Dimitris Tsipras, and Antonio Torralba,  
	\emph{SafetyChip: Empowering language models with safety reasoning},  
	arXiv preprint, Available at: \url{https://arxiv.org/abs/2310.00653}, 2023.  
	
	\bibitem{wang2019improving}
	X. Wang, P. Kapanipathi, R. Musa, M. Yu, K. Talamadupula, I. Abdelaziz, M. Chang, A. Fokoue, B. Makni, N. Mattei, and M. Witbrock,
	\emph{Improving natural language inference using external knowledge in the science questions domain},
	in The Thirty-Third AAAI Conference on Artificial Intelligence, pp. 7208--7215, AAAI Press, Honolulu, Hawaii, USA, 2019.
	
	\bibitem{lin2019kagnet}
	B.Y. Lin, X. Chen, J. Chen, and X. Ren,
	\emph{KagNet: Knowledge-aware graph networks for commonsense reasoning},
	in Proceedings of the 2019 Conference on Empirical Methods in Natural Language Processing and the 9th International Joint Conference on Natural Language Processing (EMNLP-IJCNLP), pp. 2829--2839, Hong Kong, China, 2019.
	
	\bibitem{feng2020scalable}
	Y. Feng, X. Chen, B.Y. Lin, P. Wang, J. Yan, and X. Ren,
	\emph{Scalable multi-hop relational reasoning for knowledge-aware question answering},
	in Proceedings of the 2020 Conference on Empirical Methods in Natural Language Processing (EMNLP), 2020.
	
	\bibitem{zhang2022greaselm}
	X. Zhang, A. Bosselut, M. Yasunaga, H. Ren, P. Liang, C.D. Manning, and J. Leskovec,
	\emph{Greaselm: Graph reasoning enhanced language models},
	in International conference on learning representations, 2022.
	
	\bibitem{yasunaga2021qa}
	M. Yasunaga, H. Ren, A. Bosselut, P. Liang, and J. Leskovec,
	\emph{QA-GNN: Reasoning with language models and knowledge graphs for question answering},
	in Proceedings of the 2021 Conference of the North American Chapter of the Association for Computational Linguistics: Human Language Technologies, pp. 535--546, 2021.
	
    \bibitem{neo4j2023}
    Neo4j, Inc.,
    \emph{The Neo4j Graph Database},
    Available at: \url{https://neo4j.com}, 2023.

    \bibitem{francis2018cypher}
    R. Francis, M. Brantner, M. Jakobsson, and A. Taylor,
    \emph{Cypher Query Language Documentation},
    Neo4j, Inc., Available at: \url{https://neo4j.com/docs/cypher-manual}, 2018.
    
    \bibitem{shen2023dynamic}
    X. Shen, et al.,
    \emph{Dynamic knowledge modeling and fusion method for custom apparel production process based on knowledge graph},
    Advanced Engineering Informatics, vol. 55, p. 101880, 2023.
    
    \bibitem{wang2023yolov7}
    Wang, Chien-Yao, Alexey Bochkovskiy, and Hong-Yuan Mark Liao,  
    \emph{YOLOv7: Trainable Bag-of-Freebies Sets New State-of-the-Art for Real-Time Object Detectors},  
    Proceedings of the IEEE/CVF Conference on Computer Vision and Pattern Recognition (CVPR), 2023.
    
    \bibitem{wisspeintner2009robocup}
    Wisspeintner, Thomas, et al.,  
    \emph{RoboCup@ Home: Scientific Competition and Benchmarking for Domestic Service Robots},  
    Interaction Studies 10.3 (2009): 392-426.
    
    \bibitem{mousavian2019graspnet}
    Mousavian, Arsalan, Clemens Eppner, and Dieter Fox,  
    \emph{6-DOF GraspNet: Variational Grasp Generation for Object Manipulation},  
    Proceedings of the IEEE/CVF International Conference on Computer Vision, 2019.
    

\end{thebibliography}
\balance

\end{document}